\documentclass[]{informs3} 
\usepackage{setspace}
\OneAndAHalfSpacedXI

\usepackage{float}
\usepackage{graphicx}
\floatstyle{plaintop}
\restylefloat{table}
\usepackage{tabulary}
\usepackage{multirow}
\usepackage[most]{tcolorbox}
\usepackage{multicol}
\usepackage{booktabs}
\usepackage{longtable}
\usepackage{adjustbox}
\usepackage[caption=false]{subfig}
\usepackage{fix-cm}

\usepackage[numbers]{natbib}
\bibpunct{[}{]}{,}{n}{}{,}
\def\bibfont{\small}

\usepackage{booktabs, relsize}

\TheoremsNumberedThrough     

\EquationsNumberedThrough    

\begin{document}



\RUNTITLE{Learning Based Decompostion}

\TITLE{Learning-Based Multi-Criteria Decision Making Model for Sawmill Location Problems}

\ARTICLEAUTHORS{%
\AUTHOR{Mahid Ahmed}
\AFF{School of Computing Science \& Computer Engineering, University of Southern Mississippi, \EMAIL{mahid.ahmed@usm.edu}}

\AUTHOR{Ali Dogru}
\AFF{School of Management, University of Southern Mississippi, \EMAIL{ali.dogru@usm.edu}}

\AUTHOR{Chaoyang Zhang}
\AFF{School of Computing Science \& Computer Engineering, University of Southern Mississippi, \EMAIL{chaoyang.zhang@usm.edu}}

\AUTHOR{Chao Meng}
\AFF{School of Marketing, University of Southern Mississippi, \EMAIL{chao.meng@usm.edu}}
} 

\ABSTRACT{
Strategically locating a sawmill is vital for enhancing the efficiency, profitability, and sustainability of timber supply chains. Our study proposes a Learning-Based Multi-Criteria Decision-Making (LB-MCDM) framework that integrates machine learning (ML) with GIS-based spatial location analysis via MCDM. The proposed framework provides a data-driven, unbiased, and replicable approach to assessing site suitability. We demonstrate the utility of the proposed model through a case study in Mississippi (MS). We apply five ML algorithms (Random Forest Classifier, Support Vector Classifier, XGBoost Classifier, Logistic Regression, and K-Nearest Neighbors Classifier) to identify the most suitable sawmill locations in Mississippi. Among these models, the Random Forest Classifier achieved the highest performance. We use the SHAP (SHapley Additive exPlanations) technique to determine the relative importance of each criterion, revealing the \textit{Supply-Demand Ratio}, a composite feature that reflects local market competition dynamics, as the most influential factor, followed by \textit{Road, Rail Line} and \textit{Urban Area Distance}. The validation of suitability maps generated by our LB-MCDM model suggests that 10-11\% of the MS landscape is highly suitable for sawmill location. 
}

\KEYWORDS{Multi-Criteria Decision-Making, Machine Learning, GIS-based Spatial Location Analysis, Facility Location.}

\maketitle

%

\section{Introduction} \label{sec:Introduction}

Locating industrial facility locations is a strategic decision with significant implications for operational efficiency, cost management, and long-term sustainability. This decision impacts overall supply chain performance and service delivery across various industries, including logistics \citep{logisitics_zhang2021hybrid, logistics_tirkolaee2023novel}, renewable energy \citep{energy_ahmad2022optimal}, hospitality and tourism \citep{tourism_chen2022analysis}, and manufacturing \citep{manf_liu2021sustainable}. In the wood processing industry, sawmills are the facilities that convert timber to various wood products, such as lumber and wood chips. The sawmill location selection problem presents unique complexities. For example, sawmills must be located near both forest resources and market destinations \citep{2024_Dogru}, have convenient access to roads and railways for efficient transportation \citep{2021-palander,2022-vaughan}, be positioned close to a skilled timber labor force, such as log truck drivers, logging crews, and crane operators \citep{2017-koirala}, and be situated in areas where high precipitation does not impede timber sourcing \citep{116-geisler}. A well-chosen sawmill site can reduce logistical inefficiencies, improve the use of available resources, lessen environmental impact, and contribute to the sustainable development of surrounding communities \citep{2025-khatri}. In regions where the economy depends heavily on timber production, thoughtful site selection can support local economies while maintaining ecological balance \citep{2024-jimoh}. Given the wide range of and often competing factors, sawmill site selection can be considered as a multi-criteria decision-making (MCDM) problem that requires careful joint analysis of geospatial, topographical, socio-economic, transportation, weather, operational, and market factors.

The facility location literature encompasses a variety of methodological approaches, including exact and heuristic optimization techniques \citep{arabani2012facility}, MCDM models \citep{2018_Witkowski}, GIS-based spatial analysis \citep{church2009business}, and other frameworks such as the Analytic Hierarchy Process (AHP) \citep{colak2020optimal} and Fuzzy Logic \citep{2023_Adhikari}. While GIS and MCDM methods enable the integration of diverse criteria, they often rely on subjective factor weighting based on expert opinions or researchers' judgment, which can introduce bias into site selection decisions. Conversely, exact and heuristic optimization models tend to prioritize proximity measures (i.e., transportation distance or cost), while overlooking other influential factors such as labor and resource availability, market competition, and weather. As a result, these models may produce biased or limited outputs that are partially applicable to real-world settings. This highlights an important research gap: \textit{how can we objectively determine the relative weights of multiple relevant criteria while integrating large volumes of spatial and non-spatial data from diverse sources to efficiently assess and rank the suitability of candidate locations?}

To address this research gap in MCDM, we propose a Learning-Based Multi-Criteria Decision-Making (LB-MCDM) framework that integrates Machine Learning (ML) models with GIS-based spatial analysis and MCDM to objectively incorporate a wide range of factors and predict the suitability of candidate sawmill locations by adaptively tuning the weights of these contributing factors. Unlike traditional MCDM approaches guided by expert input, our method is primarily driven by data and computation from the outset. While expert input is considered during the factor selection and model validation stages, the proposed model does not depend on subjective weighting; instead, relative factor importance is derived directly from the ML process. Utilizing real-world raster and tabular datasets, we applied LB-MCDM framework to a case study of strategic location planning for sawmills in Mississippi, one of the top timber-producing states in the U.S. \citep{2024_Dogru}. We trained five classification models, namely Random Forest Classifier (RF), Support Vector Classifier (SVC), XGBoost Classifier (XGB), Logistic Regression (LR), and K-Nearest Neighbors Classifier (KNN), using a comprehensive dataset of more than 11,000 random candidate locations, each containing values for ten relevant features, including \textit{Road Distance, Rail Line Distance, Urban Area Distance, Unemployment Rate, Terrain Slope, Market Revenue, Supply-Demand Ratio, National Land Cover} and \textit{Precipitation} and their estimated suitability scores (target) extracted from ArcGIS Pro. We validate the results of the proposed LB-MCDM framework using multiple methods, including direct comparisons with past location decisions and expert opinions. 

This research effort provides the following methodological and practical contributions. 
\begin{itemize}
    \item From a methodological point of view, we integrate ML with GIS-based spatial analysis and MCDM to minimize the reliance on subjective expert judgment and automate the industrial site location assessment process. We further improve the transparency and interpretability of the proposed LB-MCDM framework via SHAP analysis, which reveals the relative influence of each factor in determining site suitability. We also introduce a novel county-level composite feature named the \textit{Supply-Demand Ratio (SDR)}, designed to capture the trade-offs between local supply and demand when a new site is established, reflecting how existing market competition dynamics will be affected. Our findings show that \textit{SDR} consistently emerges as a highly influential factor, regardless of the ML technique applied.
    \item From a practical standpoint, our model dynamically generates a suitability map instead of a static list of candidate locations. The suitability map is continuously updated as new facilities open or existing ones close. Consequently, for any given list of candidate site locations (latitudes and longitudes), the model can automatically provide a rank-ordered list with corresponding suitability scores almost instantly. We applied the proposed LB-MCDM framework to a case study in MS using real datasets, and validated the results against the actual spatial distribution of existing sawmills and consulted with sawmill experts who make facility location decisions. The results demonstrate the robustness and applicability of the proposed methodology in real-world scenarios.
    \item Beyond serving as a practical decision support tool, the proposed framework can also assist studies that employ traditional facility location optimization models. Given that facility location problems are NP-hard, meaning that no exact algorithm is known to solve them efficiently, and the computational complexity grows exponentially with the number of candidate sites \citep{79-hakimi,83-jakob}, processing large candidate sets to identify a smaller, high-potential subset via the proposed LB-MCDM framework can significantly reduce problem size, making the proximity-based facility location models more tractable and practically relevant.
\end{itemize}  

The remainder of this paper is structured as follows. Section \ref{sec:Related_Literature} reviews the recent and relevant literature. Section \ref{sec:Methodology} introduces the LB-MCDM framework and provides an overview of key steps. Section \ref{sec:Case_Study} demonstrates the practical applicability of the proposed methodology in a case study aimed at evaluating suitable sawmill locations in Mississippi. Section \ref{sec:Discussion_Insights} discusses the managerial insights derived from the case study. Finally, Section \ref{sec:Conclusion} summarizes the key findings, discusses the limitations of the study, and suggests potential directions for future research.

\section{Related Literature} \label{sec:Related_Literature}

The literature relevant to our work lies at the intersection of four research streams: MCDM for site selection,  
GIS-based spatial analysis, ML-assisted site selection, and plant location problems. Since the literature is quite rich, we will focus on the most recent and relevant papers. Although an exhaustive review is beyond our scope, we refer interested readers to surveys on the details of facility location problems \citep{2020-celik}, multi-criteria location models \citep{2010-farahani}, and GIS-based approaches \citep{kuhaneswaran2025comprehensive}.

\subsection{Multi-Criteria Decision Making for Facility Location Problem}

MCDM has long been used to evaluate facility locations, particularly when multiple conflicting criteria must be considered. These techniques have proven effective across a range of applications, including urban infrastructure planning \citep{banaei2014maximal}, logistic center location \citep{2018_Witkowski}, and locating Photovoltaic (PV) systems in the renewable energy sector \citep{2024_deLuis-Ruiz}. For instance, \cite{banaei2014maximal} introduced the Multi-Criteria Optimal Location (MCOL) problem and developed the Maximal Reverse Skyline Query (MaxRSKY) method to identify optimal site locations in multi-dimensional space. Their approach expanded traditional proximity-based location decisions by integrating spatial diversity and preference modeling. \cite{2018_Witkowski} conducted a comparative MCDM analysis for logistics center placement in Poland, incorporating economic zones, market access, and transport connectivity. In the renewable energy sector, \cite{2024_deLuis-Ruiz} applied a weighted MCDM model using ten spatial factors to rank land suitability for PV installations, showing how GIS integration enhances spatial decision-making. Despite their practical relevance, these MCDM methods often involve subjective factor weight assignments, which may introduce bias and constrain the generalizability of site assessment decisions.

Fuzzy logic extends traditional MCDM approaches by accounting for uncertainty in criteria evaluation. For example, \cite{2023_Adhikari} used fuzzy logic and location-allocation modeling to determine optimal sites for hardwood cross-laminated timber (CLT) plants in Tennessee based on proximity to sawmills, transportation networks, and timber supply. AHP is another widely adopted MCDM extension. Examples of the use of AHP for solar PV site design and landfill selection can be found in \cite{colak2020optimal, shriki2023prioritizing} and \cite{islam2018landfill}. Similar to traditional MCDM approaches, these structured and intuitive extensions also rely on expert-defined weights and continue to face challenges related to generalizability. Our study addresses this gap by introducing an objective, data-driven, generalizable, and replicable MCDM methodology that integrates ML and GIS-based spatial location analysis to improve industrial facility location decisions.  

\subsection{GIS-based Spatial Location Analysis}
Using common factor analysis and geostatistical regression, \cite{aguilar2007factors} analyzed the spatial distribution of the softwood lumber industry in the U.S. South, identifying timber availability, supplier proximity, and vertical integration as key drivers, with spatial dependencies and clustering influenced by production efficiencies and local resources. While not explicitly focused on site selection, these findings underscore the importance of spatial patterns and resource-based advantages in sawmill placement. \cite{diaz2008making} illustrated the integration of GIS into the MCDM process to support forest management strategies. \cite{richardson2016spatial} employed GIS analysis to evaluate how sawmill locations affect the forest products industry in northern Colorado. \cite{2023_Adhikari} demonstrated the use of GIS in identifying optimal sites for hardwood CLT plants in Tennessee, highlighting three locations with substantial annual production capacity. Beyond forestry, GIS has also been utilized in site selection for wind farms \citep{bilgili2024explainability, razeghi2023multi}, landfills \citep{islam2018landfill}, and power plants \citep{colak2020optimal}. None of these studies, however, incorporates ML to objectively determine factor weights, which play a critical role in site assessment decisions.

\subsection{Machine Learning Assisted Site Selection}
Recent advances in ML offer data-driven alternatives to traditional MCDM techniques, improving the scalability and accuracy of site selection models. For example, \cite{verma2024optimal} utilized ML models to optimize distributed energy generation placement by analyzing load profiles and minimizing losses in power grids. In the renewable sector, \cite{bilgili2024explainability} used a suite of ML models such as XGB and LightGBM to evaluate wind farm sites in Türkiye based on twelve environmental, economic, and social features. Their model achieved high accuracy (XGB: 96.07\%) and is aligned well with existing wind turbine locations, underscoring ML's potential for spatial planning. Similarly, \cite{liu2023optimization} applied ML to the placement of oil wells in the petroleum industry, reporting an increase in cumulative oil production compared to traditional models. In the hospitality sector, \cite{yang2015hotel} combined ML with GIS to assess optimal hotel locations in Beijing, offering dynamic and web-based visualizations to support decision-making. \cite{2022-al-ruzouq} proposed an AHP-based framework integrated with ML algorithms to optimize the location of waste-to-energy (WTE) facilities in UAE, considering social, environmental, and economic factors. This approach achieved up to 94.6\% accuracy and identified 16.6\% of the area as highly suitable. \cite{2024-yunusoglu} presented an ML-based two-stage approach for biomass-to-bioenergy supply chain network design to locate undesirable facilities in Türkiye. These studies, however, primarily focus on other sectors like energy or urban development, with relatively limited application in timber-related industries. Our work addresses this gap by focusing specifically on the sawmill site selection problem, an economically significant yet underexplored area.

\subsection{Plant Location Models}
The plant location problem (a.k.a., the facility location problem) involves selecting the optimal locations from a set of candidate sites, which is NP-hard \citep{104-drezner}. Most plant location models in the timber industry rely on heuristics for large problem instances or exact optimization techniques for computationally tractable problem instances \citep{2005-weintraub}. For instance, \cite{117-vanzetti} presentd a MILP model to compare individual plant locations versus industrial clusters within the forest supply chain, aiming to optimize production and logistics decisions in the Argentine forestry sector. Similarly, \cite{114-duarte} proposed a MILP model for locating biofuel plants in Colombia, specifically applying the model to a second-generation bioethanol plant using coffee cut stems. Some studies focus on locating satellite wet yards, a storage area where the harvested timber is sorted and temporarily stored before being transported to sawmills. \cite{2020-aguiar} evaluated the allocation of wood storage yards in an Amazonian sustainable forest by comparing exact optimization methods with metaheuristic approaches. Similarly, \cite{2009-chan} employed MILP to determine optimal locations for satellite yards in Canadian forestry. Some studies incorporate MCDM and plant location. \cite{2023_Adhikari}, for instance, used fuzzy MCDM and location-allocation modeling to determine suitable locations for hardwood CLT manufacturing, considering the proximity to sawmills, timber supply, and transportation networks.
Our work advances plant location modeling by integrating GIS-based raster analysis and ML techniques to systematically evaluate and rank a large number of candidate sites, thereby reducing problem size and enhancing computational tractability before applying exact optimization methods.

\section{Methodology: LB-MCDM Framework} \label{sec:Methodology}
\begin{figure}[!htbp]
    \centering
    \includegraphics[width=0.975\linewidth]{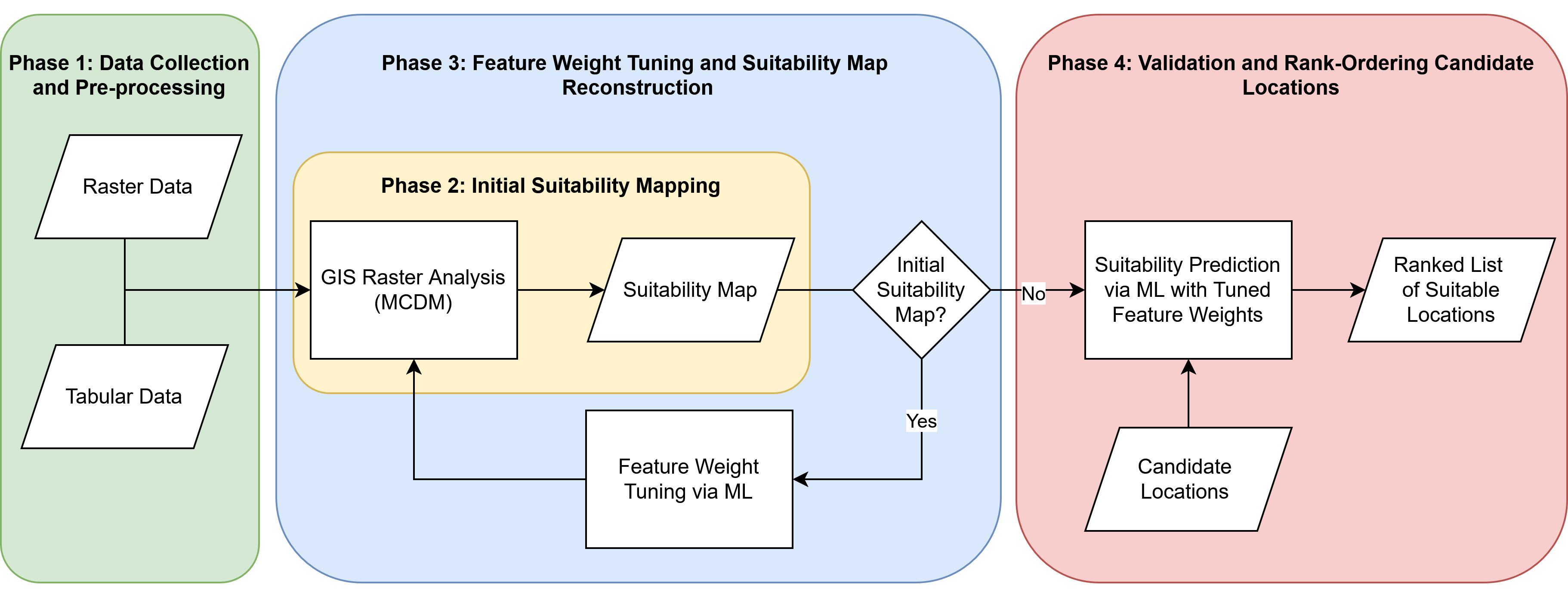}
    \caption{LB-MCDM Process Flow}
    \label{fig:processflow}
\end{figure}

This section outlines the key components of the proposed LB-MCDM framework, which integrates GIS-based geospatial analysis with ML to address limitations in traditional MCDM approaches. Consider a decision maker who seeks to select a subset of promising site locations denoted by $\mathcal{V}^s$ from the set of $N$ candidate locations $\mathcal{V}^c=\{v_1,v_2,\dots,v_N\}$ such that $\mathcal{V}^s\subset \mathcal{V}^c$. 
The decision maker evaluates these candidate locations using $\mathcal{K}$ criteria, each criterion $i\in \mathcal{K}$ represented by a raster layer. A raster layer $\textbf{R}$ is a two-dimensional matrix with $m$ rows and $n$ columns, where each cell $r_{xy}$ contains a value of the spatial feature at location $(x,y)$. Given a vector of raster layers $\vec{X}=(\textbf{R}_1, \textbf{R}_2, \dots, \textbf{R}_K)$ and corresponding vector of weights $\vec{w}=(w_1, w_2, \dots, w_K)$, the weighted sum layer $\hat{\textbf{R}}(\vec{X},\vec{w})$ is computed as:
\begin{align}
   \hat{\textbf{R}}(\vec{X},\vec{w}) &= \sum_{i=1}^{K} w_i \textbf{R}_i 
\end{align}
In practice, these weights are often determined subjectively, based on expert opinion survey results, empirical findings in the literature, or researchers' own assessment. To reduce subjectivity and potential bias, we propose employing ML techniques. The proposed LB-MCDM framework consists of four key phases: 1) data collection and pre-processing, 2) initial suitability mapping, 3) feature weight tuning and suitability map reconstruction, and 4) validation and rank-ordering of candidate locations. The process flow is summarized in Figure \ref{fig:processflow}. In the following subsections, we elaborate on the key phases.
\subsection{Data Collection and Preprocessing} In this phase, decision-makers gather both raster and tabular data relevant to the facility location problem under investigation. In the sawmill location problem domain, raster data, such as land cover, terrain slope, proximity to roads and railways, and proximity to urban areas, represent spatially continuous variables derived from geospatial layers. In contrast, tabular data, including labor statistics, timber supply and demand, local market revenue, and precipitation, provide structured and attribute-based information. Once the relevant data is collected, tabular data is converted into a raster format to enable spatial processing within a GIS environment. This ensures consistency in data format and facilitates seamless integration during subsequent analysis. Data preprocessing is an essential step because raw data is often inconsistent or incomplete, which can negatively impact the ML model's performance. After collecting the data, it is essential to clean the datasets, handle missing values, perform necessary transformations, and scale the features. 
\subsection{Initial Suitability Mapping} With the raster pre-processed, a weighted sum overlay is performed in the GIS system to generate an initial suitability map. At this stage, equal weights are assigned to each of the $K$ features to avoid premature bias $(w_i=\dfrac{1}{K}, \forall i \in \mathcal{K})$. The weighted sum layer $\hat{\textbf{R}}(\vec{X},\vec{w})$ provides location-specific suitability scores visually represented through a color gradient. The decision maker then samples a large number of random points from the initial suitability map, each with associated predictor values (features) and a target to be used in training ML classifiers.
\subsection{Feature Weight Tuning and Suitability Map Reconstruction} ML phase follows a structured process that includes data partitioning (training and testing), feature selection, model development, hyperparameter tuning, and model evaluation. Since multiple ML classifiers can be used in the proposed framework, let $\vec{y}_j=(y_1, y_2, \dots,y_N)\in \{0,1,2,3\}^N$ denote the multi-class target variable (i.e. 0: not suitable, 1: somewhat suitable, 2: suitable, 3: highly suitable) for the ML classifier $j \in \mathcal{J}$. Each classifier $j$ predicts $\vec{y}_j$ based on the vector of sample candidate location input features $\vec{x}$ derived from $\vec{X}$ and the initial vector of equal weights $\vec{w}$ such that $\vec{y}_j=f_j(\vec{x}, \vec{w})$ and yields $\vec{\alpha}_j=(\hat{\alpha}_1, \hat{\alpha}_2, ..., \hat{\alpha}_N)\in [0,1]^N$, the resulting vector of the likelihoods of candidate locations in $\mathcal{V}^c$ being suitable. Then, the initial feature weight vector $\vec{w}$ is updated using the SHAP values computed for feature $i$ by ML classifier $j$ such that $w_{ij}=\phi_{ij}, \forall i \in \mathcal{K} \text{ and } \forall j \in \mathcal{J}$, resulting in the updated feature weight vector denoted by $\vec{w}^{'}$. These tuned weights are re-integrated into the GIS system to reconstruct the suitability map $\hat{\textbf{R}}(\vec{X},\vec{w}^{'})$, reflecting the updated factor importance. The decision-makers then resample from this refined map to generate a new dataset, which is used to retrain the ML classifiers. Then employing standard model evaluation techniques, the decision-maker identifies the best-performing ML classifier out of $J$ classifiers denoted by $\vec{y}^{*}=f^{*}(\vec{x}, \vec{w}^{'})$ with likelihoods scores $\vec{\alpha}^*=(\hat{\alpha}_1^{*}, \hat{\alpha}_2^{*}, ..., \hat{\alpha}_N^{*})\in [0,1]^N$.
\subsection{Validation and Rank-ordering Candidate Locations} After retraining, the best-performing ML classifier's final predictions are validated using existing site locations and expert opinions. Once validated, the classifier can assess and rank candidate sites, enabling data-driven prioritization based on predicted suitability. The candidate locations are first sorted in descending order of their estimated suitability scores: $\mathcal{V}^c=\{v_{(1)}, v_{(2)}, ...,v_{(N)}\}$ such that $\hat{\alpha}_{(1)}^{*}\geq \hat{\alpha}_{(2)}^{*}\geq...\geq \hat{\alpha}_{(N)}^{*}$. Then, the decision maker chooses $M$ best suitable of $N$ candidate locations: $\mathcal{V}^s=\{v_{(m)}\in \mathcal{V}^c|m\leq M\}$.

\section{Case Study: Sawmill Location in Mississippi via LB-MCDM} \label{sec:Case_Study}

To demonstrate the applicability and effectiveness of the proposed LB-MCDM framework, we conducted a case study focusing on the state of Mississippi (MS). The state of MS is one of the leading timber-producing states in the U.S., with forests covering 63\% of its land equivaling to about 19 million acres of forestland \citep{MDAC}. Timber ranks as the state’s third-largest commodity, supporting approximately 50,000 jobs and supplying over 140 sawmills \citep{2024_Dogru}. These factors make MS a strong candidate for studying the sawmill location assessment problem. 

\subsection{Data Collection and Pre-processing}

We identified ten key criteria influencing sawmill site suitability, guided by insights from prior literature, consultations with policymakers, industrial relevance, and the available data. Table~\ref{tab:data_summary} summarizes the data sources, measurements, and intended purposes for each feature. We use these datasets to construct the suitability layer in ArcGIS software and train ML classifiers. Each criterion (feature) is briefly described below to contextualize its role in LB-MCDM framework:

\begin{table}[!ht]
\begin{footnotesize}
\setlength{\extrarowheight}{1pt}
\tabcolsep=0.2cm 
    \centering
    
    \begin{tabular}{>{\raggedright}p{2cm}>{\raggedright}p{2cm}>{\raggedright}p{6.5cm}p{4cm}}
         \textbf{Data} & \textbf{Source} & \textbf{Measurement} & \textbf{Purpose} \\\toprule
         National Land Cover Data & US Census Bureau & \textit{Highly suitable:} Developed areas located on barren land with open space and low intensity, \newline \textit{Suitable:} Forest land and grassland, \newline \textit{Somewhat suitable:} Developed areas located on shrubland and cultivated land with medium and high intensity, \newline \textit{Not suitable:} Water space, and wetlands. & To distinguish candidate locations based on their land suitability and potential for sustainable development.\\\midrule
         Transportation Data& USGS National Transportation Dataset & \textit{Highly suitable:} 0m $<$ Value $\leq$ 500m, \newline \textit{Suitable:} 500m $<$ Value $\leq$ 1000m, \newline \textit{Somewhat suitable:} 1000m $<$ Value $\leq$ 2000m, \textit{Not suitable:} Value $>$ 2000m. \newline \textit{Source:} \cite{2024_deLuis-Ruiz} & To measure proximity to major roads and rail lines in order to improve transportation access and reduce logistical costs. \\\midrule
         Slope Data & USGS open topography &  \textit{Highly suitable:} $0 \leq \text{Slope} \leq 7\%$, \newline \textit{Suitable:} $8 \leq \text{Slope} \leq 13\%$, \newline \textit{Somewhat suitable:} $14 \leq \text{Slope} \leq 20\%$, \newline \textit{Not suitable:} $21 \leq \text{Slope} \leq 90\%$. \newline \textit{Source:} \cite{2022_Susiati} & To determine the slope of potential sites to ensure ease of access and support more efficient transportation and construction. \\\midrule
         Urban Data & US Census Bureau & \textit{Highly suitable:} Value $< 10{,}000m$, \newline \textit{Suitable:} $10{,}000m \leq \text{Value} < 20{,}000m$, \newline \textit{Somewhat suitable:} $20{,}000m \leq \text{Value} < 50{,}000m$, \newline \textit{Not suitable:} $\text{Value} \geq 50{,}000m$ & To ensure better access to utilities and services such as electricity, gas, water, communication, banking, and healthcare facilities. \\\midrule
         Labor Data & U.S. Bureau of Labor Statistics & County-level unemployment rate for labor whose age is greater than 16. & To assess the availability of labor in the area to support sawmill operations. \\\midrule
         Timber Demand Data & Forisk Mill Capacity Database& The data provides detailed information on sawmills' timber processing capacities and demands in tons. & To align site selection with existing timber demand and operational capacity constraints. \\\midrule
         Timber Supply Data & MS Dept. of Agriculture \& Commerce  & Total available timber volume in tons at the county level. & To prioritize locations with sufficient timber availability. \\\midrule
         Revenue Data & Nexis Uni & The annual average revenue of wood manufacturing companies in USD. & To estimate the local timber market demand. \\\midrule
         Precipitation Data & National Center for Environmental Information  & County-level rainfall in inches in 2024.  & To prioritize areas with lower precipitation levels for operational efficiency, as significant precipitation can hinder timber harvesting and transportation activities. \\\midrule
         
    \end{tabular}
    \caption{Summary of Datasets}
    \label{tab:data_summary} 
    \end{footnotesize}
\end{table}

\begin{itemize}
    \item \textbf{National Land Cover Data (NLCD):} Sourced from the U.S. Census Bureau, the NLCD layer classifies land into usability categories. \textit{Highly suitable} areas include barren land, open space, and low-intensity development. \textit{Suitable} areas consist of forest land and grassland, while \textit{somewhat suitable} areas encompass developed zones situated on shrubland and cultivated land with medium to high development intensity. Areas classified as \textit{not suitable} include water bodies, wetlands, and national forests. The dataset allows us to distinguish candidate site locations based on their relative suitability and potential for sustainable development \citep{2023_Adhikari,2022_Susiati}. 
    \item \textbf{Transportation Data:} The U.S. Geological Survey's National Transportation Dataset enables us to measure proximity to major roads and rail lines. Proximity to roads and rail lines reduces inbound and outbound logistics costs for sawmills.
    \item \textbf{Slope Data:} Open topography data from the U.S. Geological Survey allows the assessment of terrain slope, classifying potential sites from flat to very steep slopes \citep{2022_Susiati}. This raster data helps identify land that supports easy construction and transportation routes. Avoiding steep areas reduces development costs and improves transportation efficiency. 
    \item \textbf{Urban Data:} Proximity to urban areas improves access to utilities such as electricity, gas, water, and communication, which are critical for sawmill operations, and increases access to services like banking and healthcare, which are important for the sawmill, its employees, and customers. This raster dataset identifies urban zones to ensure infrastructure support and enhance access to essential services \citep{2023_Adhikari}.
    \item \textbf{Labor Data:} 
Labor force availability is vital for sustaining sawmill operations, with sites near urban areas preferred due to better access to skilled workers. We assess the current labor supply in MS using Local Area Unemployment Statistics (LAUS) as defined by the U.S. Bureau of Labor Statistics. The tabular dataset allows for retrieving the county-level unemployment rate for the labor population in 82 MS counties aged 16 and older. A higher unemployment rate may indicate greater labor availability, making the location more favorable for sawmill operations.
    \item \textbf{Timber Demand Data:} Forisk 2024 Mill Capacity Database provides the timber processing capacities, timber demands, and geolocations of more than 140 wood-consuming mills in MS. We use this raster data to align site selection with existing market demand and capacities of competitor sawmills. 
    \item \textbf{Timber Supply Data:} Provided by the MS Department of Agriculture and Commerce, this county-level tabular dataset provides available timber in tons in 82 MS counties. A higher timber supply indicates greater suitability for establishing a potential sawmill. 
    \item \textbf{Market Revenue Data:} We retrieved revenue data from Nexis Uni for Mississippi companies that depend on lumber and wood products, excluding existing sawmills, in tabular format. To evaluate local market potential, we calculated the total revenue of these companies within a 75-mile radius of each sawmill. The 75-mile rule is a widely used rule of thumb in the timber industry, reflecting the typical procurement or sales radius \citep{2007_nate_anderson}. It balances transportation costs and supply chain efficiency, making it a practical benchmark for siting decisions. Proximity to markets offers significant operational advantages to sawmills, including increased sales opportunities, faster delivery times, lower outbound transportation costs, and a smaller carbon footprint \citep{2024_Dogru}. 
    \item \textbf{Precipitation:} We obtained county-level precipitation data (in inches) for all 82 Mississippi counties from the year 2024, sourced from the National Centers for Environmental Information. Precipitation refers to any form of water, such as rain, sleet, or snow, that falls from the atmosphere to the ground, typically measured in inches. High precipitation negatively impacts upstream timber supply chains by reducing accessibility for logging, damaging forest roads, and causing safety hazards and transportation delays \citep{2024_Dogru}. Additionally, wet logs are prone to quality degradation, increased drying costs, and operational inefficiencies. 
    \item\textbf{Supply-Demand Ratio (SDR):}
To capture market competition dynamics, we propose a county-level composite metric, supply-demand ratio, which reflects the trade-offs between timber supply and demand following the introduction of a new sawmill in a county. A higher SDR indicates a more favorable location for a new sawmill, as it reflects greater timber availability and lower competition for timber in the area.

\begin{definition}
    Assume there are $J$ counties, indexed by $j\in \mathcal{J}$. Let $\mathcal{N}_j$ and $S_j$ denote the set of existing sawmills and the available timber supply in tons in county $j$, respectively. In addition, let $D_{ij}$ be the annual tons of timber demanded by the existing sawmill $i$ from county $j$. Now, suppose that we open a new sawmill in county $j$, which will demand $D_{new,j}$ tons of timber annually. Assuming the supply remains constant, the supply-demand ratio for county $j$, denoted by $SDR_j$, is calculated using the following expression:
    \begin{align}
        SDR_j&=\dfrac{S_j}{\displaystyle\sum_{i\in \mathcal{N}_j}D_{ij}+D_{new,j}}, \quad \forall j \in \mathcal{J}
    \end{align}
\end{definition}

To calculate $D_{ij}$, we employ the 75-mile radius rule: each sawmill $i$ sources timber within a 75-mile radius. Since this radius may span multiple counties, we allocate the demand proportionally based on the portion of the radius area that lies within each county so that the total demand by sawmill $i$ can be found $D_i=\displaystyle\sum_{j\in\mathcal{J}}D_{ij}, \: \forall i \in \mathcal{N}_j$. Figure \ref{fig:SDR} compares county-level timber supply, timber demand, and the calculated SDRs. Note that SDR is not a simple, static supply-to-demand ratio for each county; rather, it captures the dynamic reallocation of sawmill-level timber demand using the 75-mile radius rule after hypothetically placing a new sawmill in each county.
\end{itemize}

\begin{figure}
    \centering
    \includegraphics[width=0.3\linewidth]{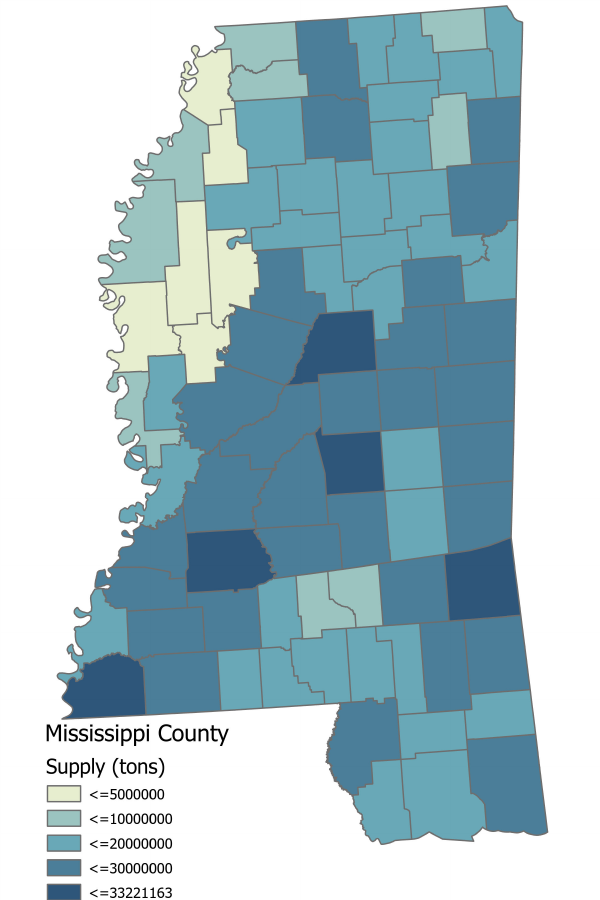}
    \includegraphics[width=0.3\linewidth]{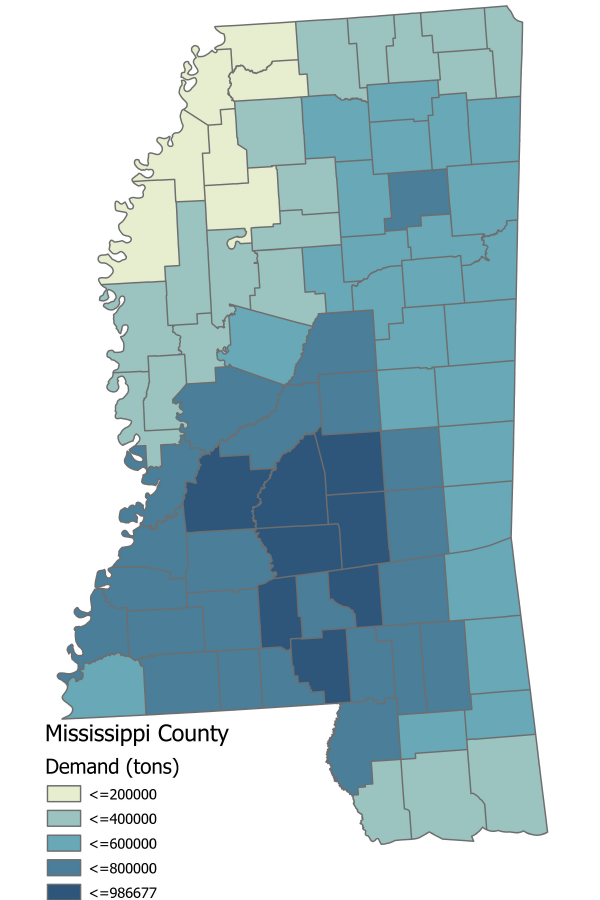}
    \includegraphics[width=0.3\linewidth]{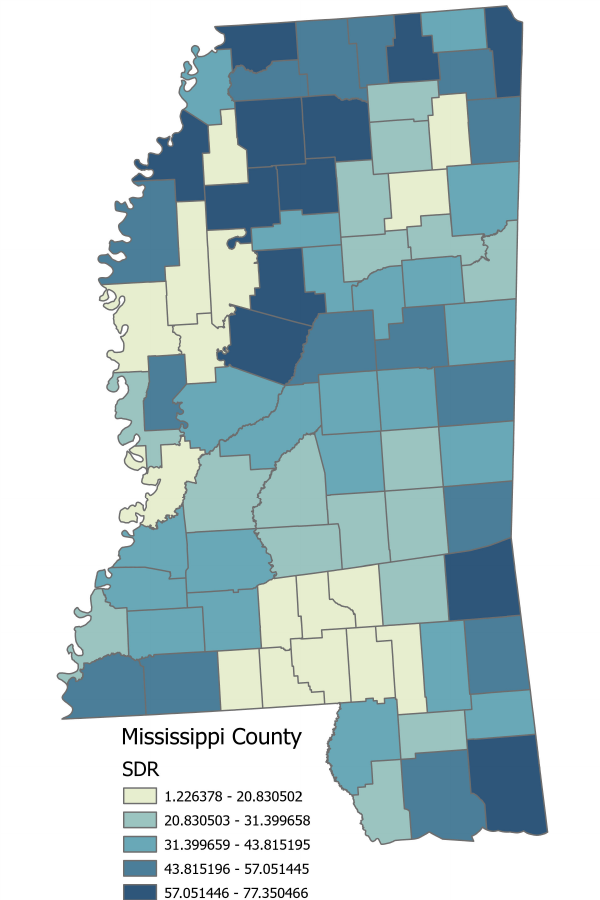}
    \caption{Comparison of Timber Supply, Timber Demand, and Supply Demand Ratio in MS}
    \label{fig:SDR}
\end{figure}


During data preprocessing, we cleaned the datasets by removing records with critical missing values and imputing remaining gaps using the statistical mode to preserve distributional properties. After cleaning, we applied \textit{standard scaling} to all features, transforming them to have a mean of 0 and a standard deviation of 1 to ensure equal contribution and fair weight computation across variables. Finally, we converted tabular datasets (labor, timber demand, timber supply, revenue, and precipitation datasets) into a raster format to enable spatial processing in the ArcGIS Pro software.
\subsection{Initial Suitability Mapping} Initially, we assigned equal weights to all nine features (i.e., $w_i=\dfrac{1}{9}=0.111, \:\forall i \in \mathcal{K}$) and calculated the weighted sum layer $\hat{\textbf{R}}$, which serves as our initial suitability map. From this map, we initially sampled 10,000 random points (locations), each containing values for nine predictor variables and the corresponding estimated suitability score, which is a non-negative continuous variable. We classified these random locations into four suitability levels: \textit{highly suitable, suitable, somewhat suitable}, and \textit{not suitable}, following the classifications adopted in prior studies. To determine thresholds, we calculated the range of suitability scores of these random points and divided the range into four equal intervals, where the highest suitability interval indicates \textit{highly suitable} and the lowest interval \textit{not suitable}. After observing significant class imbalances among the suitability categories, we drew an additional random sample of 1,467 locations to achieve a more balanced dataset. The final dataset, consisting of 11,467 observations, was reclassified into the same four categories, which mitigated but did not fully eliminate the imbalance issue. To further address this, we decided to employ synthetic minority oversampling techniques during model training. We split the resulting dataset into training (80\%) and testing (20\%) subsets.
\subsection{Feature Weight Tuning and Suitability Map Reconstruction}
At this stage, we have a large dataset to train multiple ML classifiers, consisting of nine features: \textit{Road Distance, Rail Line Distance, Urban Area Distance, Unemployment Rate, Terrain Slope, Market Revenue, Supply-Demand Ratio, National Land Cover} and \textit{Precipitation}. Our target variable \textit{Suitability} is a categorical variable with four categories: \textit{Highly Suitable, Suitable, Somewhat Suitable} and \textit{Not Suitable}. In the following subsections, we discuss how we addressed data imbalances, selected features, evaluated models, and fine-tuned feature weights using the SHAP values. We provide the details regarding the employed ML classifiers, calculation of performance metrics, and SHAP calculations in Appendices A, B, and C, respectively. 
\subsubsection{Handling imbalances in the dataset}
Despite our efforts to incorporate additional random samples, our initial suitability map revealed an unbalanced distribution across the four suitability categories, with 1,306 sites classified as highly suitable, 2,397 as suitable, 6,152 as somewhat suitable, and 1,612 as not suitable. As is widely known, imbalanced datasets can bias ML models toward the majority class, leading to misleading accuracy and poor precision and recall for underrepresented classes. To mitigate this issue, we employed the Synthetic Minority Over-sampling Technique - Edited Nearest Neighbors (SMOTE-ENN), a hybrid of SMOTE and ENN on the training data only to avoid data leakage (all other preprocessing steps were conducted separately on the training and testing datasets to ensure the validity of model evaluation). SMOTE generates synthetic samples for the minority class by interpolating between existing minority class instances and their \( k \)-nearest neighbors \citep{zhu2017synthetic}. Following the SMOTE step, ENN removes samples that are misclassified by their nearest neighbors, which helps eliminate noisy data and better preserve the class boundaries. This approach reduces duplication and improves model generalization by enhancing the quality of the synthetic samples while maintaining class separability.
\subsubsection{Feature selection} 
Feature selection aims to systematically identify and retain the most relevant features that significantly enhance ML model performance by eliminating less important, correlated, or redundant features, thereby improving accuracy, reducing overfitting, and decreasing computational complexity. In this study, we employed a sequential approach combining correlation analysis, multicollinearity testing, feature importance, and SHAP values to identify the most suitable features for our ML model.

\begin{figure}[!ht]
    \centering
    \includegraphics[width=0.75\linewidth]{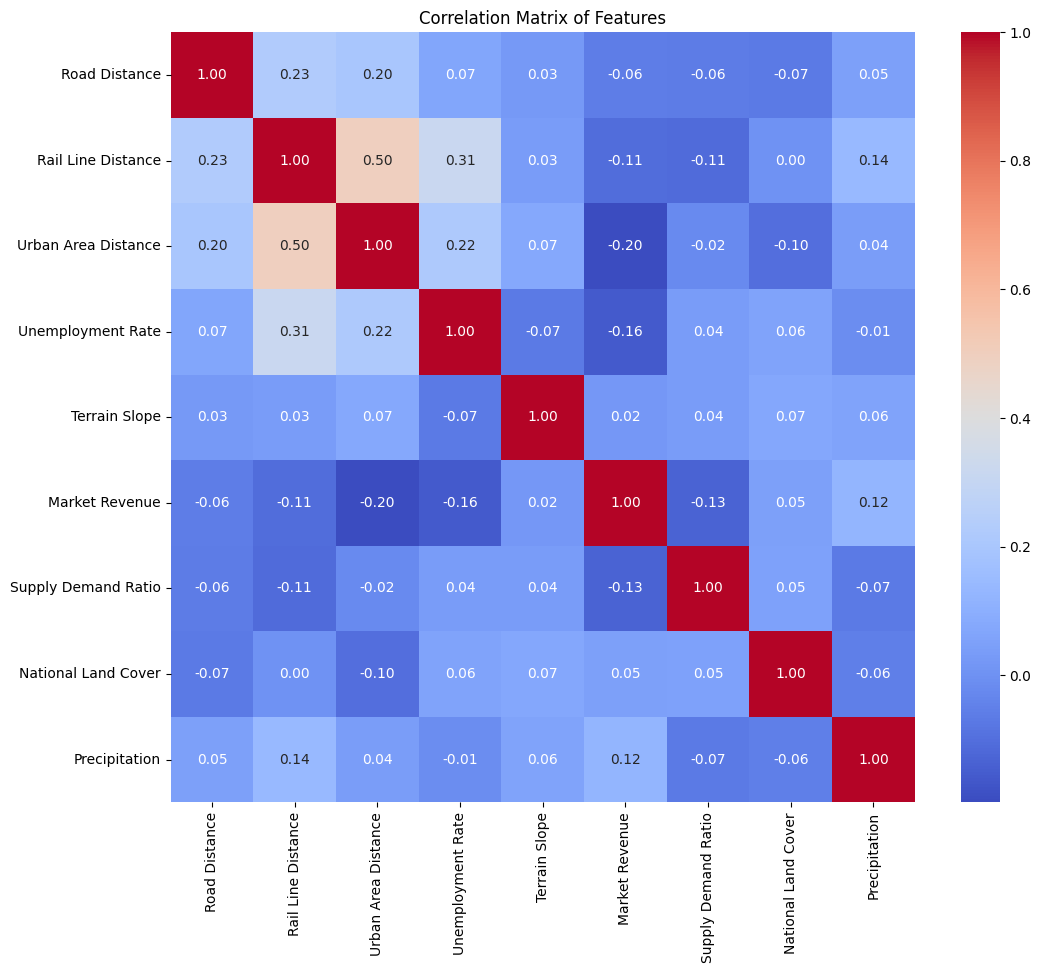}
    \caption{Correlation Matrix among the 9 input features}
    \label{fig:feature_selection}
\end{figure}

We first constructed the correlation matrix of all nine features presented in Figure \ref{fig:feature_selection} to identify highly correlated feature pairs (\(r > 0.7\) or \(r < -0.7\)), which could potentially introduce redundancy into our model. As shown in the figure, the \textit{urban area distance} and \textit{rail distance} pair have the highest correlation $(r=0.50)$, which is well below the threshold $(r=0.7)$, so we decided to retain all nine features. 

\begin{table}[!ht]
\centering
\caption{Variance Inflation Factor (VIF) and Tolerance}
\begin{tabular}{lcc}
\toprule
\textbf{Feature} & \textbf{VIF} & \textbf{TOL} \\
\midrule
Road Distance         & 1.0762 & 0.9292 \\
Rail Line Distance         & 1.4911 & 0.6706 \\
Urban Area Distance   & 1.4112 & 0.7082 \\
Unemployment Rate     & 1.1471 & 0.8718 \\
Terrain Slope                 & 1.0271 & 0.9736 \\
Market Revenue        & 1.0976 & 0.9111\\
Supply Demand Ratio   & 1.0481 & 0.9541 \\
National Land Cover   & 1.0429 & 0.9589 \\
Precipitation  & 1.0507 & 0.9518 \\
\bottomrule
\end{tabular}
\label{tab:vif_tol}
\end{table}


To further assess multicollinearity and ensure feature independence, we calculated the Variance Inflation Factor (VIF) for each feature. Although many machine learning algorithms can handle correlated inputs, assessing VIF enhances model interpretability and minimizes redundancy that could distort feature importance rankings. VIF quantifies how much the variance of a regression coefficient is inflated due to linear relationships with other variables \citep{tay2017correlation}. A VIF close to 1 suggests negligible multicollinearity; values exceeding 10 imply high multicollinearity that could affect model performance. Tolerance indicates how much of the variability in a feature is independent of other features; values closer to 0 suggest stronger dependence. As shown in Table \ref{tab:vif_tol}, all VIF scores are below 1.50 and all tolerance values exceed 0.67, indicating that there is no multicollinearity problem.

\begin{figure}[!htbp]
    \centering
    \includegraphics[width=0.48\linewidth]{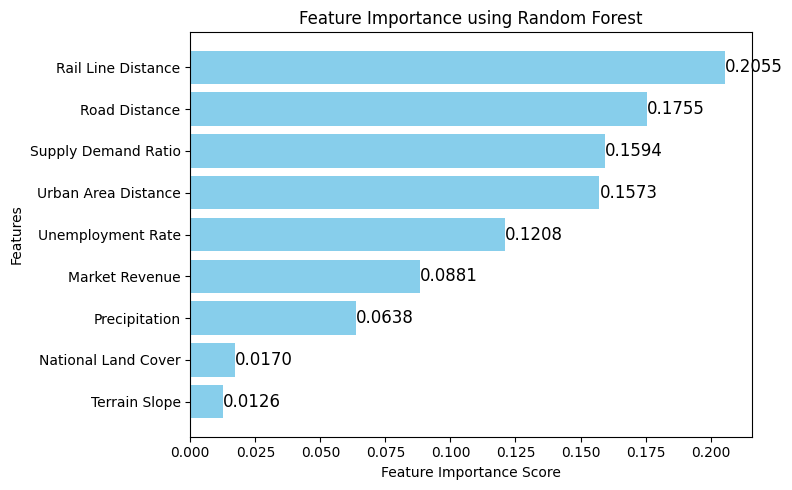}
    \includegraphics[width=0.48\linewidth]{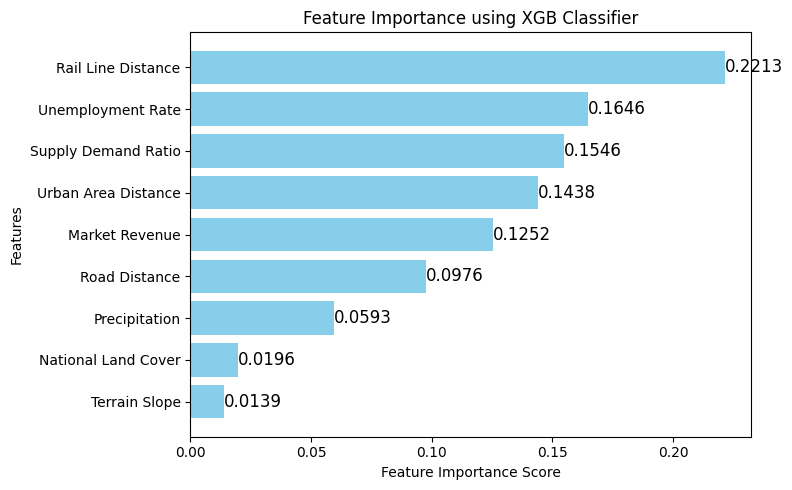}

    \includegraphics[width=0.48\linewidth]{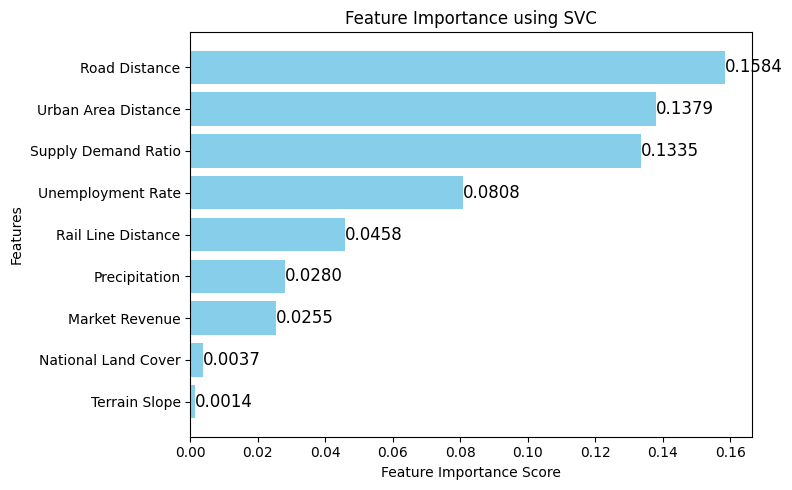}
    \includegraphics[width=0.48\linewidth]{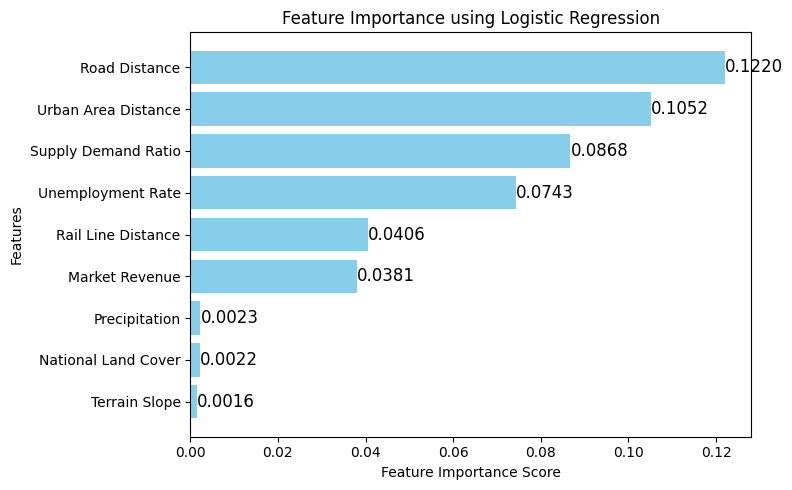}

    \includegraphics[width=0.48\linewidth]{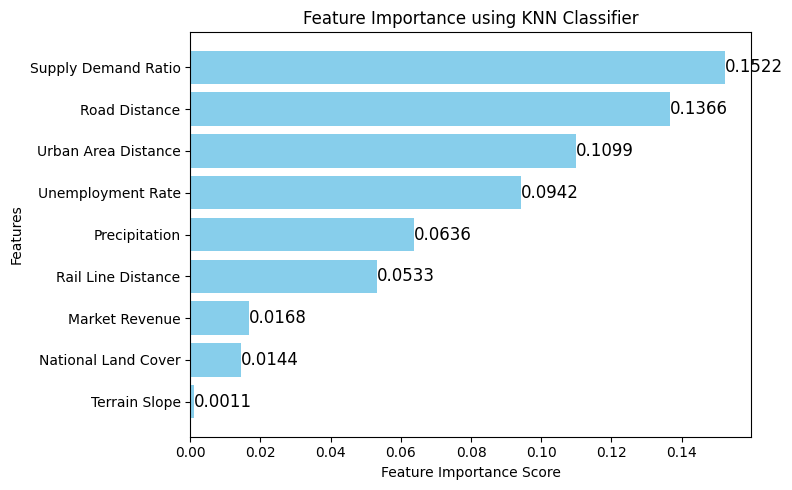}
    
    \caption{Feature importance Scores by five ML Classifiers}
    \label{fig:feature_importances}
\end{figure}

Next, we analyzed the feature importance scores of five ML models, namely RF, XGB, SVC, LR, and KNN. Figure~\ref{fig:feature_importances} presents the feature importance rankings derived from multiple ML models. The results consistently indicate that \textit{Terrain Slope} and \textit{National Land Cover} have minimal influence across all models. This finding can be attributed to the relatively flat topography of Mississippi, where slope variations are negligible across the state. Furthermore, Mississippi's extensive forest coverage reduces the variability in land cover types, thereby diminishing the significance of the National Land Cover feature in determining suitable locations for sawmill development. However, it is important to note that in more mountainous geographies with limited timber resources, these two features may play a more critical role. Due to their relatively low importance, we decided to remove \textit{Terrain Slope} and \textit{National Land Cover} and continue our analysis with the remaining seven features (\textit{SDR, Road Distance, Rail Line Distance, Urban Area Distance, Unemployment Rate, Market Revenue}, and \textit{Precipitation}).

\subsubsection{Model Evaluation and Classification Performance}

\begin{figure}[!htbp]
    \centering
    \includegraphics[width=0.4\linewidth]{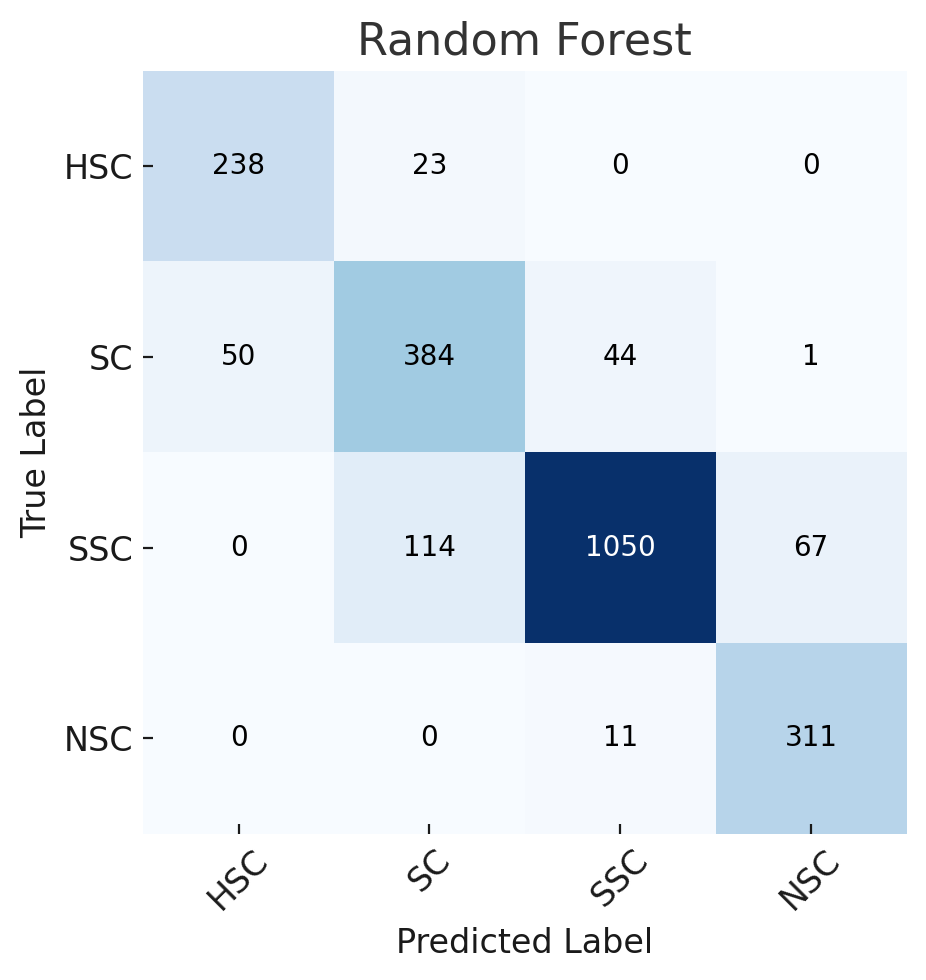}
    \includegraphics[width=0.4\linewidth]{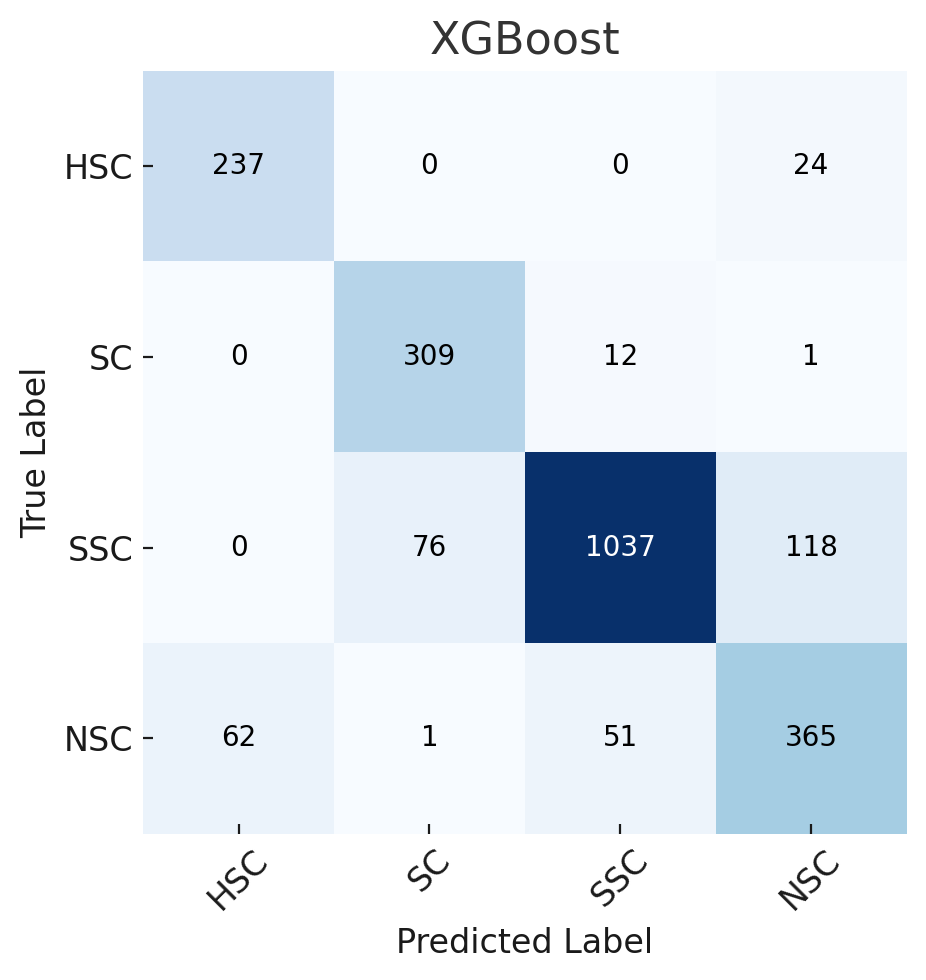}

    \includegraphics[width=0.4\linewidth]{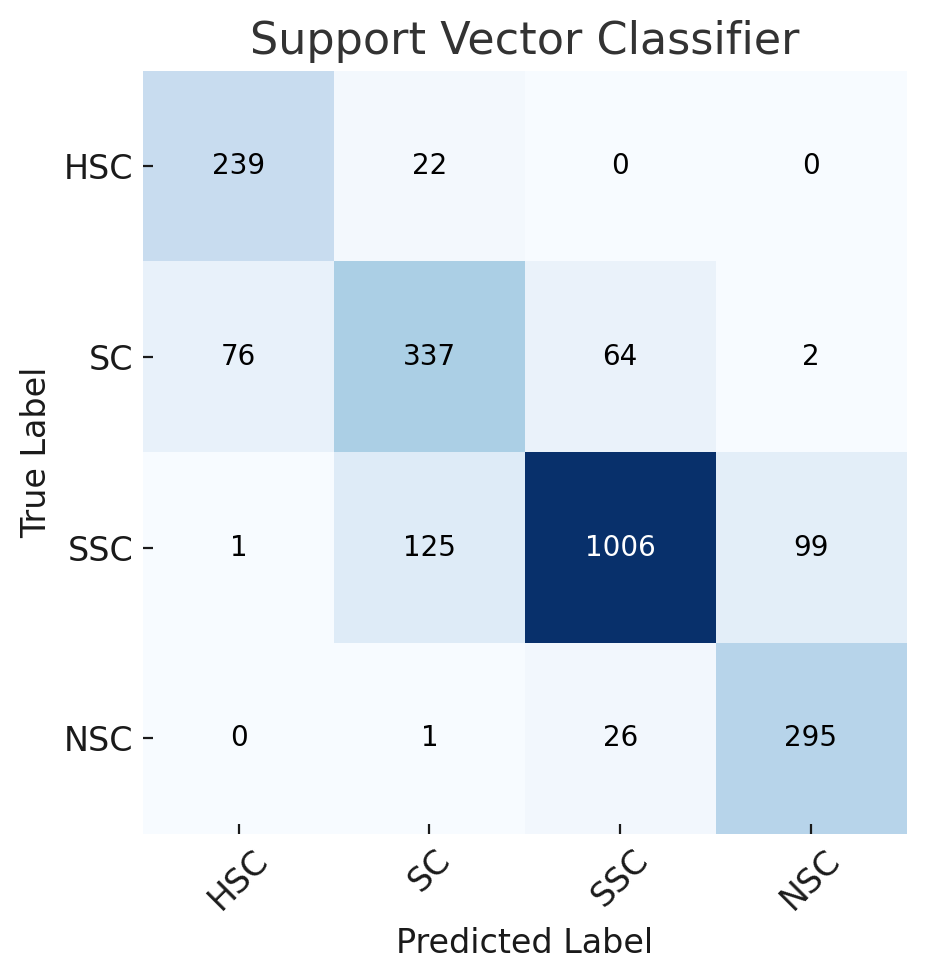}
    \includegraphics[width=0.4\linewidth]{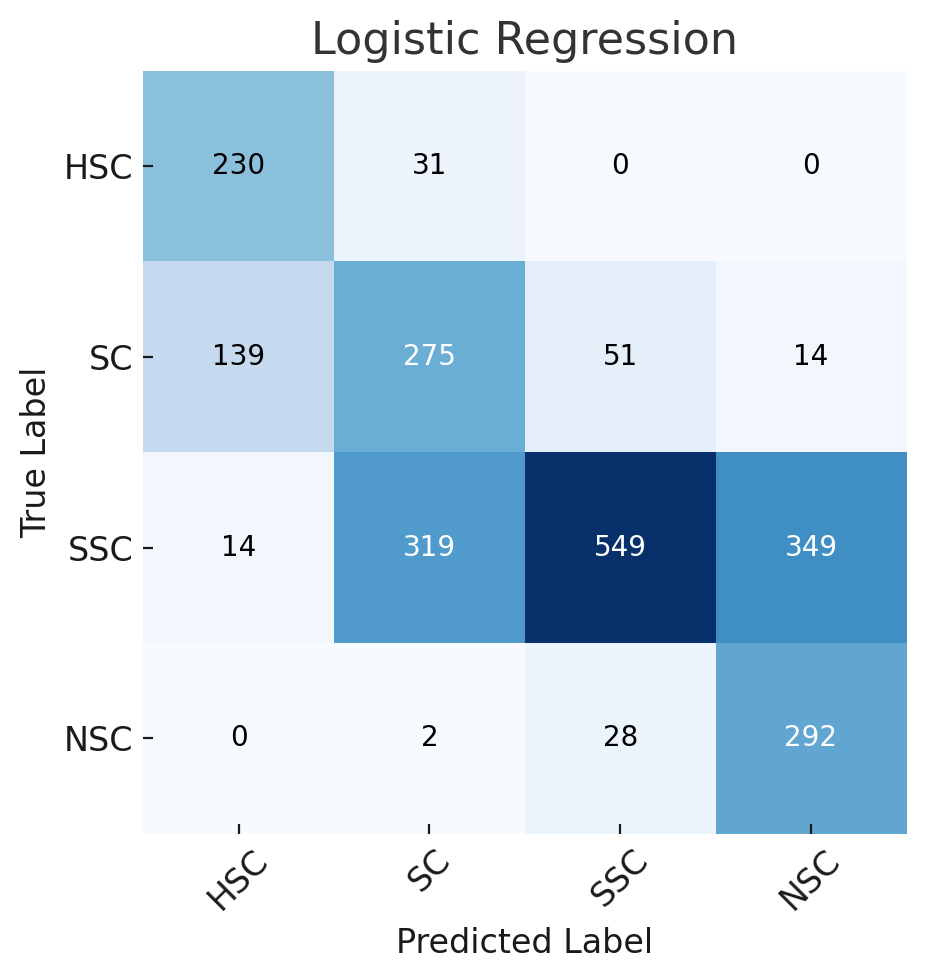}

    \includegraphics[width=0.4\linewidth]{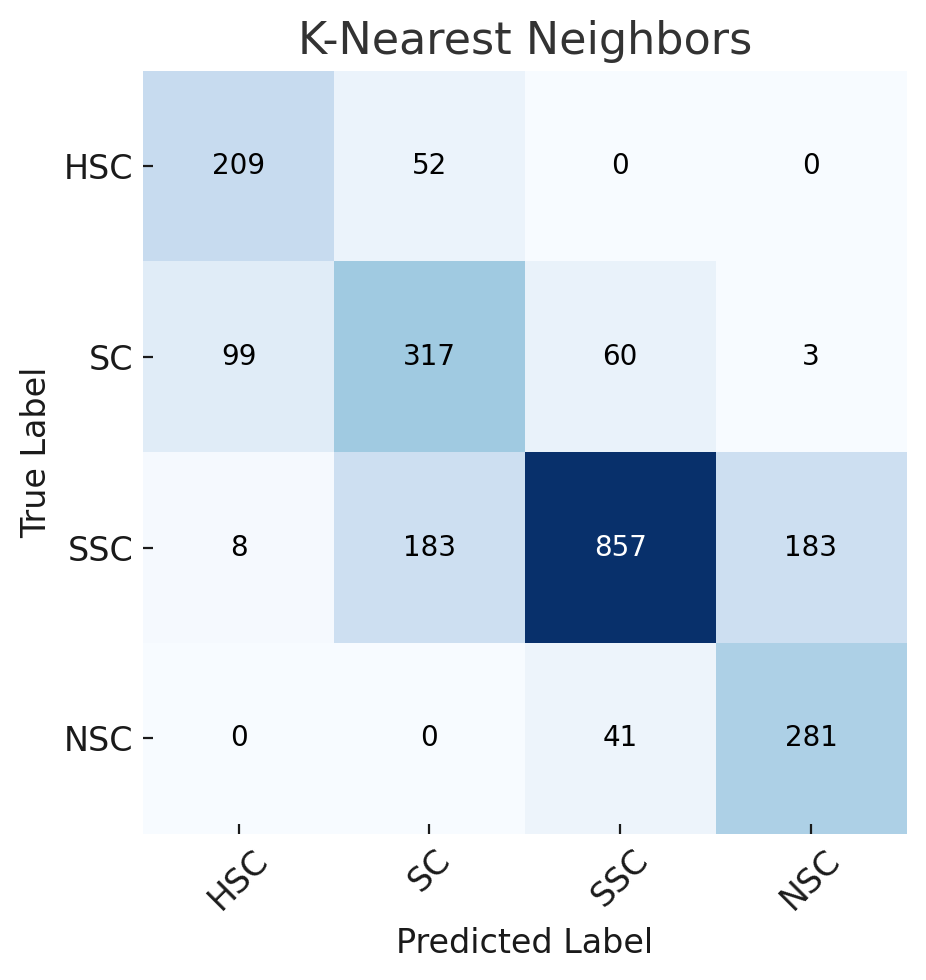}
    
    \caption{Confusion Matrix of five ML Classifiers}
    \label{CM}
\end{figure}

\begin{figure}[!htbp]
    \centering
    \includegraphics[width=0.48\linewidth]{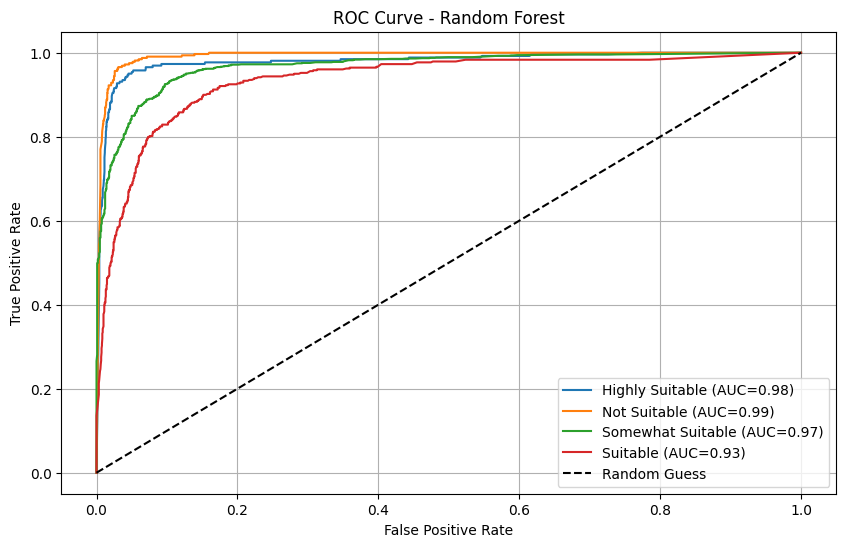}
    \includegraphics[width=0.48\linewidth]{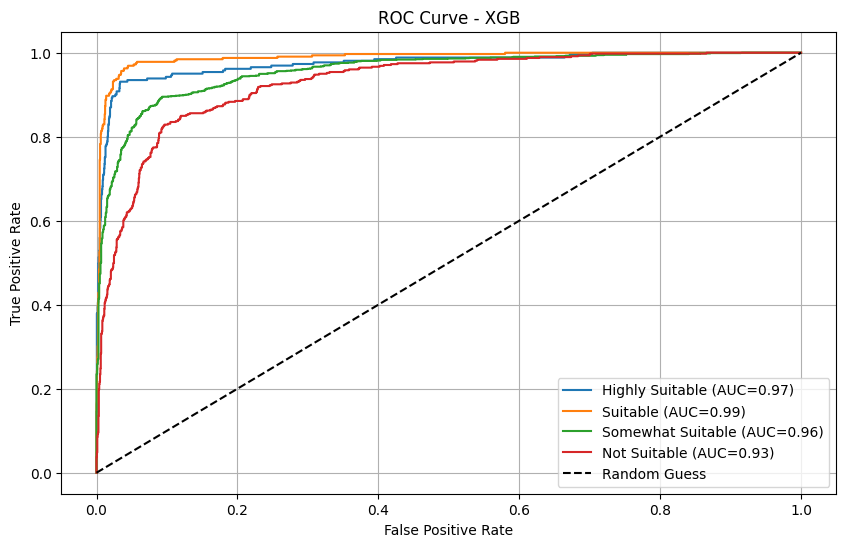}

    \includegraphics[width=0.48\linewidth]{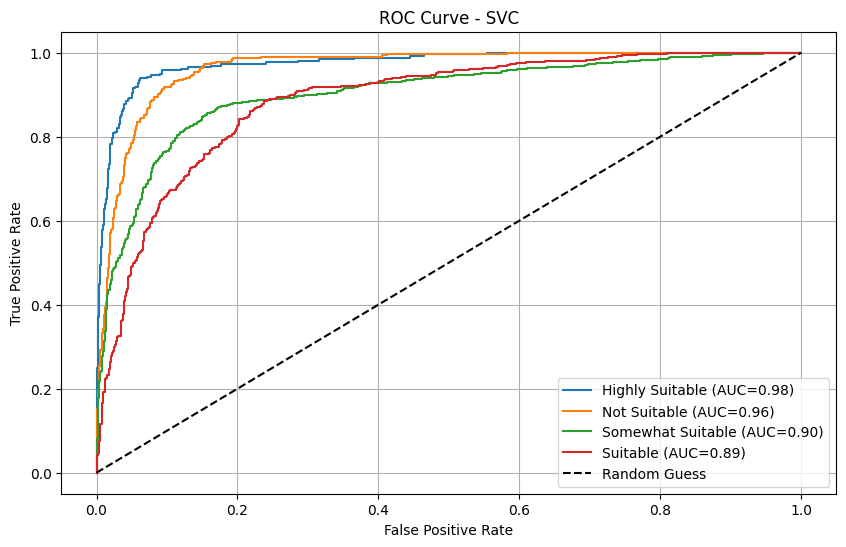}
    \includegraphics[width=0.48\linewidth]{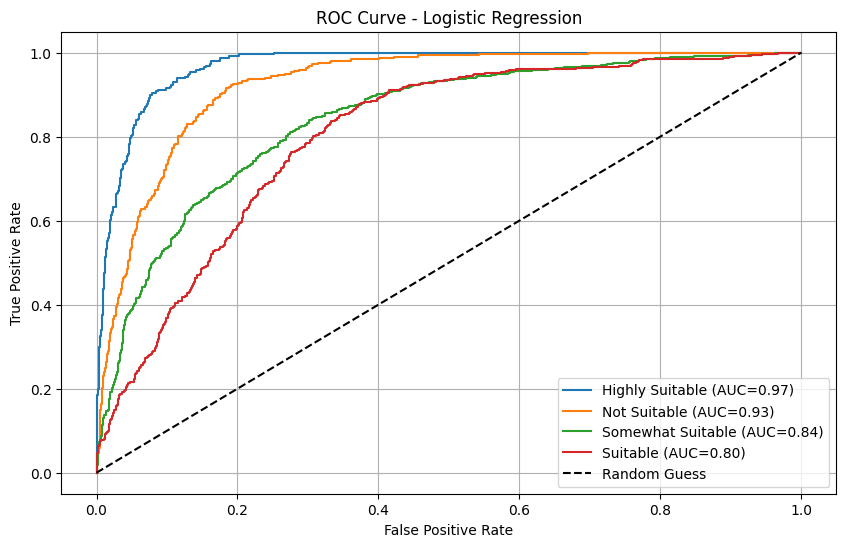}

    \includegraphics[width=0.48\linewidth]{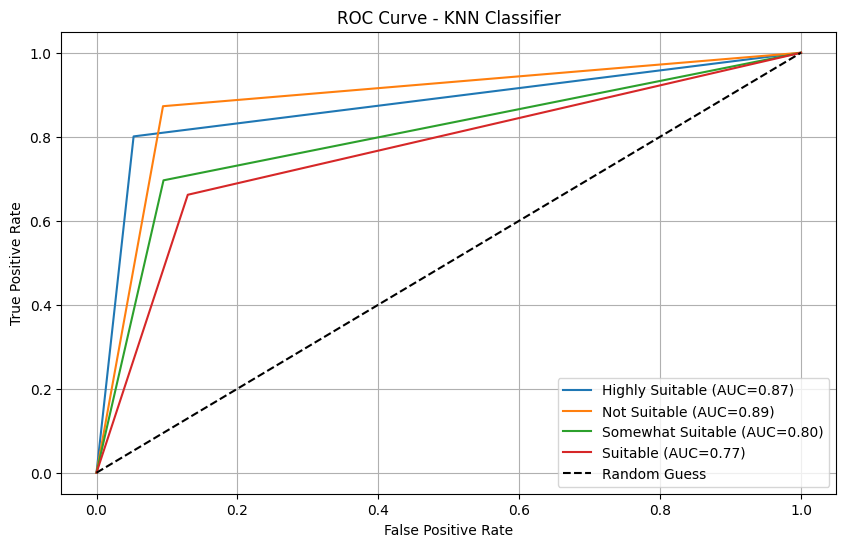}
    
    \caption{AUCROC Curve of five ML Classifiers}
    \label{AUCROC}
\end{figure}

Using selected seven features, we trained and evaluated five ML classifiers: RF, XGB, SVC, LR, and KNN. We optimized the parameters of each ML model through hyperparameter tuning. Performance metrics are presented in Table~\ref{tab:classifier_performance}, with confusion matrices shown in Figure~\ref{CM} and AUC-ROC curves in Figure~\ref{AUCROC}.

\begin{table}[!htbp]
\centering
\footnotesize
\begin{tabular}{lcccccccccc}
\hline
\textbf{Classifier} & \textbf{Accuracy} & \textbf{Recall} & \textbf{Precision} & \textbf{F1 Score} & \textbf{HSC} & \textbf{SC} & \textbf{SSC} & \textbf{NSC} & \textbf{AUC}  \\
\hline
\textbf{RF} & \textbf{0.8648} & \textbf{0.8648} & \textbf{0.8734} & \textbf{0.8663} & 0.9119 & 0.8017 & \textbf{0.8530} & \textbf{0.9658} & \textbf{0.9656} \\
\textbf{XGBoost} & 0.8495 & 0.8495 & 0.8588 & 0.8511 & 0.9080 & \textbf{0.9596} & 0.8424 & 0.7620 & 0.9569 \\
\textbf{SVC} & 0.8186 & 0.8186 & 0.8286 & 0.8199 & \textbf{0.9157} & 0.7035 & 0.8172 & 0.9161 & 0.9159 \\
\textbf{LR} & 0.5870 & 0.5870 & 0.6919 & 0.5862 & 0.8812 & 0.5741 & 0.4460 & 0.9068 & 0.8597 \\
\textbf{KNN} & 0.7257 & 0.7257 & 0.7600 & 0.7313 & 0.8008 & 0.6618 & 0.6962 & 0.8727 & 0.8142 \\
\hline
\end{tabular}
\caption{Performance comparison. HSC: Highly Suitable Class, SC: Suitable Class, SSC: Somewhat Suitable Class, NSC: Not Suitable Class.\newline}
\label{tab:classifier_performance}
\end{table}

Among all models, the RF classifier demonstrated the highest overall performance, achieving an accuracy of 0.8648, a precision of 0.8734, and the highest AUC score of 0.9656.RF also achieved the highest per-class accuracy in the \textit{Somewhat Suitable Class} (0.8530) and the \textit{Not Suitable Class} (0.9658), while demonstrating strong performance across the remaining classes.

XGB, SVC, and KNN also delivered a decent performance. XGB slightly outperformed SVC overall, achieving an accuracy of 0.8495 and an AUC of 0.9569. It also produced the best classification accuracy for the Suitable Class (0.9596) and performed competitively in the Highly Suitable Class (0.9080). SVC achieved an accuracy of 0.8186, a precision of 0.8286, and an AUC of 0.9159. It showed strength in classifying extreme categories, with the highest accuracy for the \textit{Highly Suitable Class} (0.9157). KNN, while lower in overall metrics, reached a respectable accuracy of 0.7257 and achieved the best performance in the Not Suitable Class (0.8727), indicating potential utility in conservative classification contexts.
In contrast, LR exhibited the weakest overall performance, with an accuracy of 0.5870, F1 score of 0.5862, and the lowest AUC score of 0.8597. Its poor per-class accuracy, particularly 0.4460 for the Somewhat Suitable Class, limits its effectiveness for this multi-class classification task. The AUC-ROC curves in Figure~\ref{AUCROC} further support these findings, confirming the superior class separation capabilities of RF, XGB, and SVC, compared to the relatively weaker performance of LR and KNN.

Table \ref{tab:feature_comparison} demonstrates that removing two features from the model did not lead to any significant degradation in performance. In fact, the Random Forest (RF) classifier with seven features achieved comparable, and in several cases slightly superior, results to the nine-feature version. The seven-feature RF model maintained higher accuracy, recall, precision, and F1 score, while the AUC values (0.9629 for nine features and 0.9656 for seven features) remained nearly identical, indicating consistent discriminative capability. Similarly, per-class accuracies showed only minor variations without any meaningful decline in classification quality. These findings suggest that the reduced seven-feature RF model offers a more efficient and streamlined configuration without compromising accuracy or predictive robustness.

\begin{table}[!htbp]
\centering
\footnotesize
\begin{tabular}{lccccccccc}
\toprule
\textbf{Model} & \textbf{Accuracy} & \textbf{Recall} & \textbf{Precision} & \textbf{F1 Score} & \textbf{HSC} & \textbf{SC} & \textbf{SSC} & \textbf{NSC} & \textbf{AUC} \\
\midrule
\textbf{RF (9 Features)} & 0.8404 & 0.8404 & 0.8529 & 0.8430 & 0.8621 & 0.7912 & 0.8278 & 0.9441 & 0.9629 \\
\textbf{RF (7 Features)} & 0.8648 & 0.8648 & 0.8734 & 0.8663 & 0.9119 & 0.8017 & 0.8530 & 0.9658 & 0.9656 \\
\bottomrule
\end{tabular}
\caption{Comparison of Model Performance with 9 and 7 Features using the RF model}
\label{tab:feature_comparison}
\end{table}

\subsubsection{Ensuring Explainability via SHAP Analysis}
Once the models are trained and their performances evaluated, the next step is to ensure explainability by quantifying the influence of each feature on the model's predictions. While gain-based feature importance metrics can aid feature selection, they often lack consistency and completeness in explaining predictions \citep{lundberg2018consistent}. SHAP analysis, grounded in cooperative Game Theory and first introduced by Shapley \citep{shapley1953stochastic}, addresses this gap by fairly distributing the prediction output among the contributing features. SHAP provides both local and global interpretability through sample-wise feature attributions, typically visualized using beeswarm and bar plots.

\begin{figure}[!htbp]
    \centering
    \includegraphics[width=0.48\linewidth]{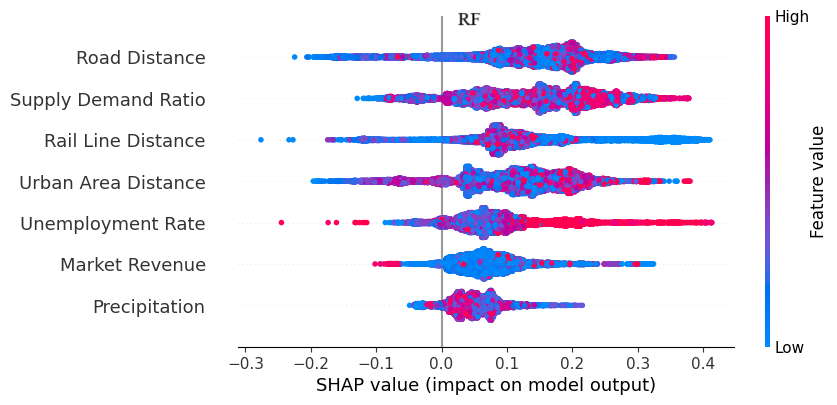}
    \includegraphics[width=0.48\linewidth]{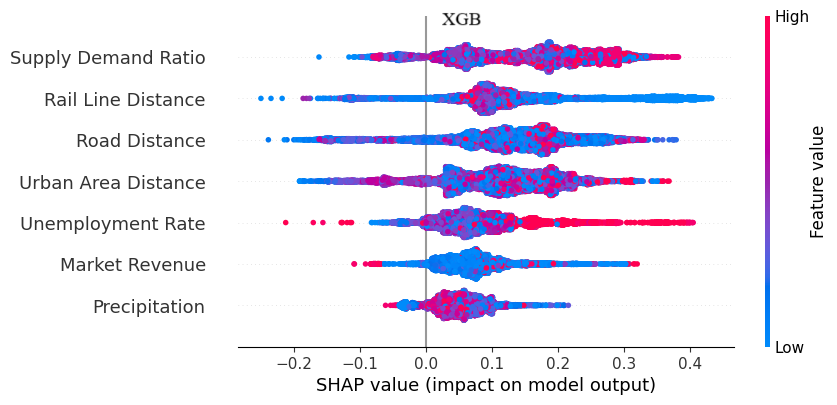}

    \includegraphics[width=0.48\linewidth]{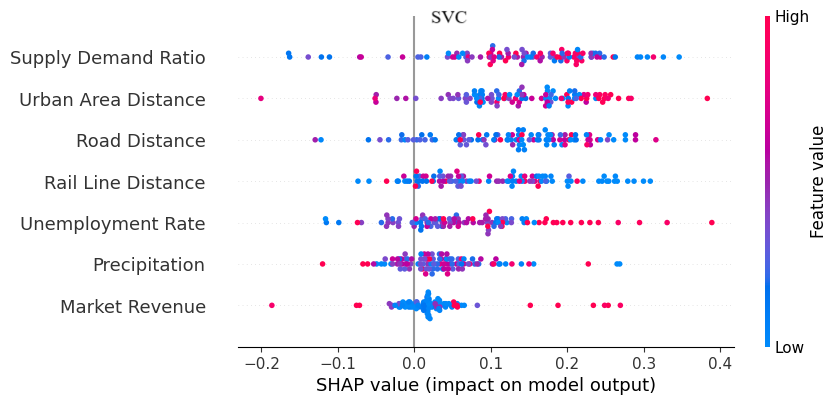}
    \includegraphics[width=0.48\linewidth]{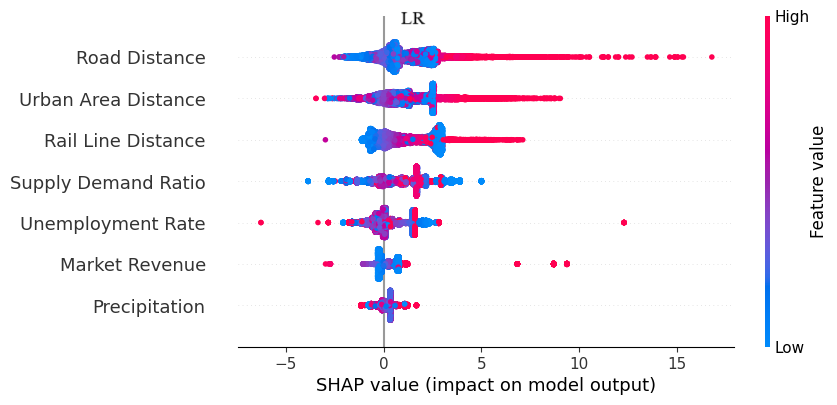}

    \includegraphics[width=0.48\linewidth]{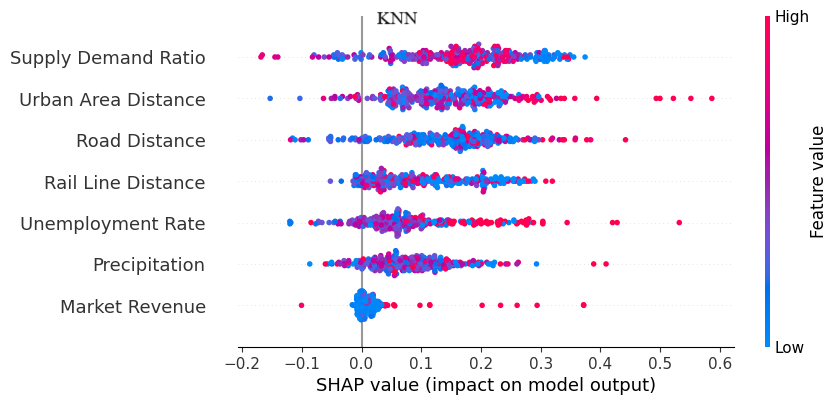}
    
    \caption{Impact of individual features on the output of five machine learning classifiers (RF, XGB, SVC, LR, and KNN) using SHAP analysis. Each plot visualizes the contribution of features to the model predictions. The color scale represents the feature value (red for high and blue for low), while the SHAP values on the x-axis indicate the magnitude and direction of each feature’s influence on the model output.}

    \label{impact}
\end{figure}

\begin{figure}[!htbp]
    \centering
    \includegraphics[width=0.48\linewidth]{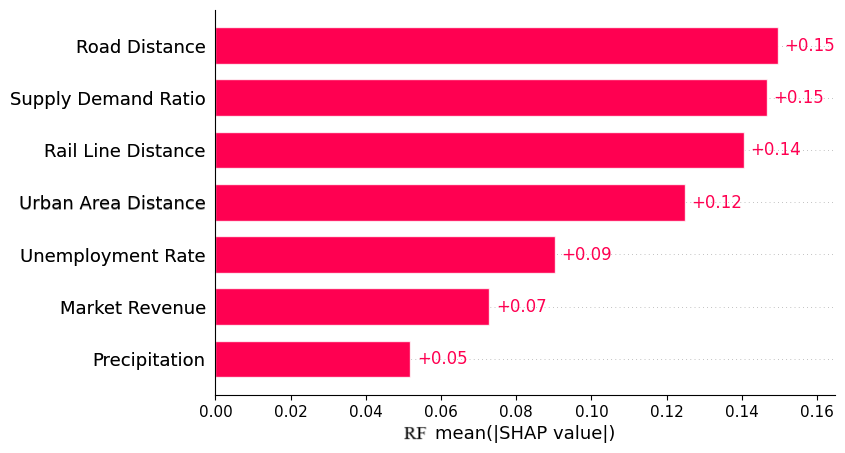}
    \includegraphics[width=0.48\linewidth]{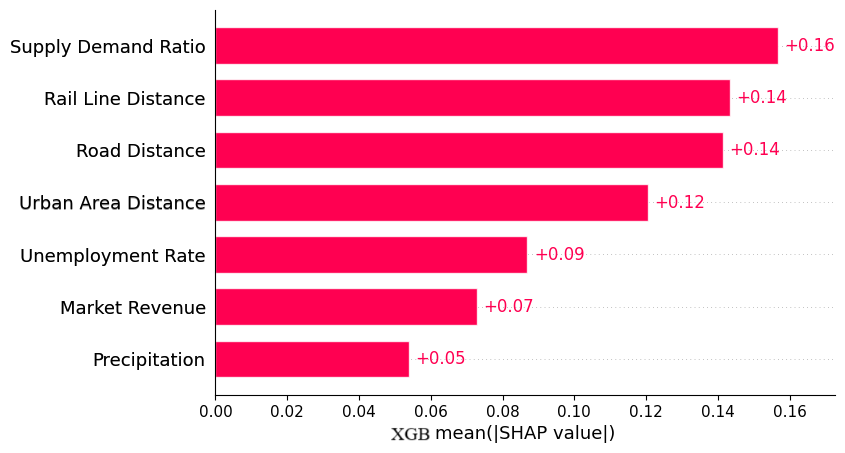}

    \includegraphics[width=0.48\linewidth]{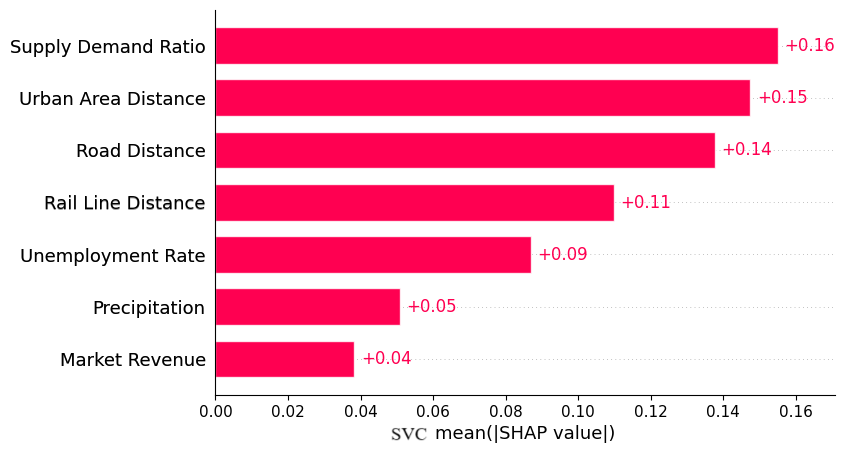}
    \includegraphics[width=0.48\linewidth]{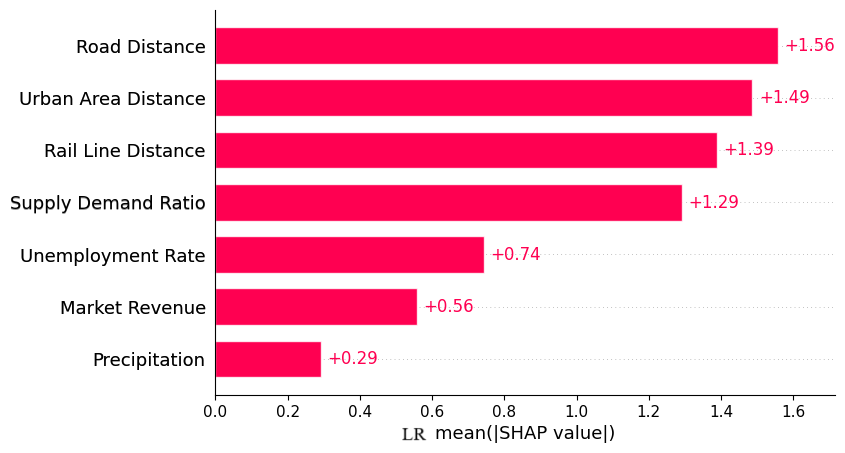}

    \includegraphics[width=0.48\linewidth]{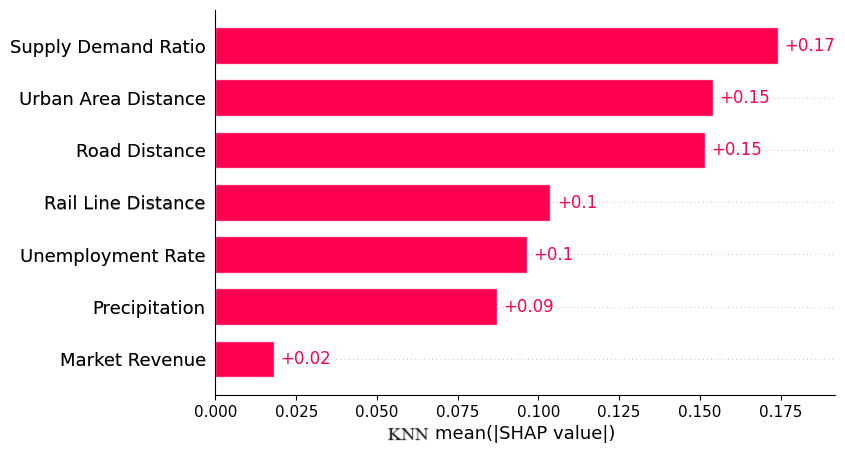}
    
    \caption{Mean SHAP values of the top features across five machine learning classifiers (RF, XGB, SVC, LR, and KNN). The horizontal bars represent the average absolute SHAP values, quantifying each feature’s overall contribution to model predictions.}
    \label{mean_shap}
\end{figure}

Figure~\ref{impact} shows that the SDR has a strong positive impact on model predictions, making it the most influential feature for the RF, XGB, SVC, and KNN models. This finding underscores the significance of market competition and resource availability in sawmill location decisions. As a composite construct combining timber supply and demand, SDR effectively reduces feature dimensionality while enhancing prediction accuracy, illustrating how meaningful feature engineering can improve model performance. As can be seen in the figure, the \textit{Road Distance} appears as the most significant feature in the Logistic Regression model. Features such as \textit{Precipitation} and \textit{Market Revenue} consistently exhibit the lowest importance across all models and contribute only marginally and positively to the predictions.

Figure~\ref{mean_shap} shows the mean SHAP values across the five ML models used. It confirms that \textit{SDR}, \textit{Road Distance}, \textit{Urban Area Distance}, and \textit{Rail Line Distance} are consistently among the most important predictors across all models. On the other hand, \textit{Market Revenue} and \textit{Precipitation} exhibit relatively lower average SHAP values, highlighting their relatively limited contribution to the sawmill suitability assessment.

By averaging the SHAP values across the five models, we observe that \textit{SDR} holds the highest overall importance, followed by \textit{Road Distance} and \textit{Urban Area Distance}. These findings strengthen the justification for prioritizing these criteria in sawmill site suitability modeling.
\subsubsection{Suitability Map Reconstruction with Tuned Weights}
Once the SHAP values are retrieved, they can be used to fine-tune the weights of the raster layers on ArcGIS Pro software. Then the raster analysis is run once more to generate the corresponding final suitability maps. For each ML model, a separate suitability map was developed by incorporating SHAP-based feature contribution scores. Unlike the initial suitability map, where all features were assigned equal weights, the final suitability maps incorporated feature weights computed from SHAP values, reflecting their relative importance in the model's decision-making process (see Table \ref{tab:comadjweights}, $baseline\: weight=1/7=0.143$). The resulting suitability scores ranged from 0 to 1 and were classified into four categories using the Jenks Natural Breaks method in ArcGIS Pro software. Following classification, the spatial distribution of each suitability class was calculated for all models to compare performance and spatial agreement across the predictions.

\begin{table}[htbp]
\centering
\caption{Comparison of Adjusted Weights}
\label{tab:comadjweights}
\footnotesize
\begin{tabular}{>{\raggedright}p{1.5cm}>{\raggedright}p{1.5cm}>{\raggedright}p{1.5cm}>{\raggedright}p{1.5cm}>{\raggedright}p{1.5cm}>{\raggedright}p{1.5cm}>{\raggedright}p{1.5cm}p{1.5cm}}
\toprule
\textbf{Adjusted Weights} & \textbf{Road Dist.} & \textbf{Rail Line Dist.} & \textbf{Urban Dist.} & \textbf{Unemp. Rate} & \textbf{Market Rev.} & \textbf{SDR} & \textbf{Precipit.} \\
\toprule
\textit{Baseline}& 0.143 & 0.143 & 0.143 & 0.143 & 0.143 & 0.143 & 0.143 \\\midrule
RF      & 0.195 & 0.182 & 0.156 & 0.117 & 0.091 & 0.195 & 0.065 \\
XGB      & 0.182 & 0.182 & 0.156 & 0.117 & 0.091 & 0.208 & 0.065 \\
SVC      & 0.189 & 0.149 & 0.203 & 0.122 & 0.054 & 0.216 & 0.068 \\
LR       & 0.213 & 0.190 & 0.204 & 0.101 & 0.077 & 0.176 & 0.040 \\
KNN      & 0.192 & 0.128 & 0.192 & 0.128 & 0.026 & 0.218 & 0.115 \\
\bottomrule
\end{tabular}%
\end{table}

\begin{figure}[!htbp]
    \centering
    \includegraphics[width=0.975\linewidth]{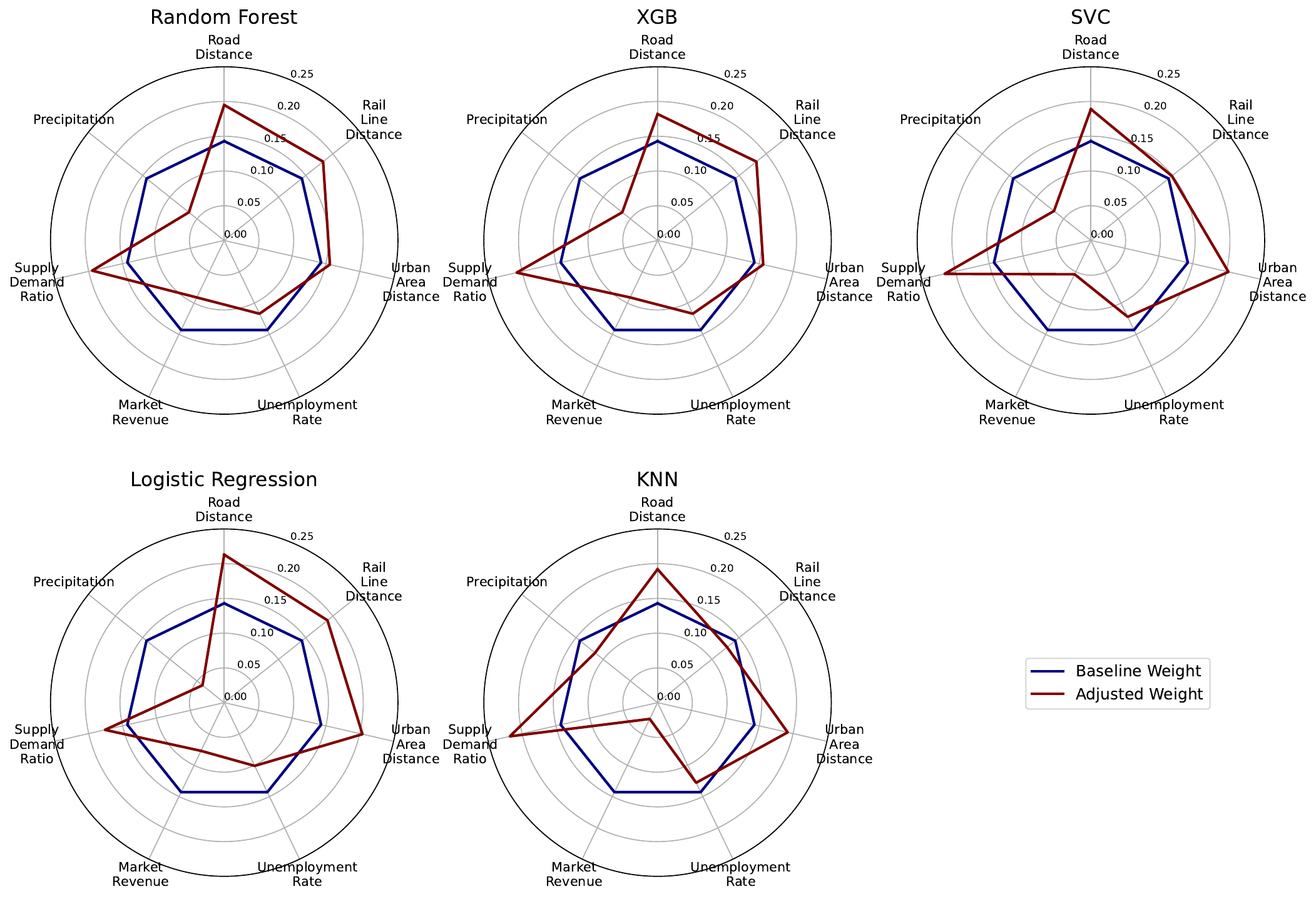}
    \caption{Comparison of Adjusted Weights by Different ML Classifiers}
    \label{fig:radarchart}
\end{figure}

Figure \ref{fig:radarchart} compares the adjusted weights of the seven relevant features across the five ML models, highlighting \textit{SDR} and the three proximity features (\textit{Road Distance}, \textit{Rail Line Distance}, and \textit{Urban Area Distance}) as the most significant factors influencing sawmill location suitability.

\subsection{Validation and Rank-Ordering Candidate Locations}\label{sec:suitability_map}
\begin{figure}[!ht]
    \centering
    \includegraphics[width=0.32\linewidth]{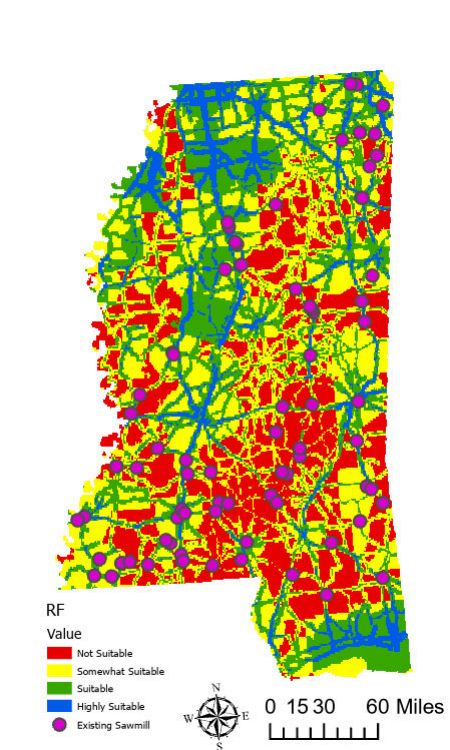}
    \includegraphics[width=0.32\linewidth]{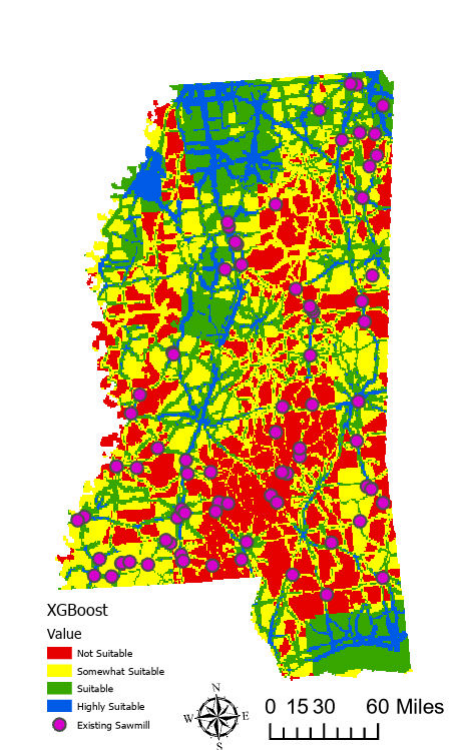}

    \includegraphics[width=0.32\linewidth]{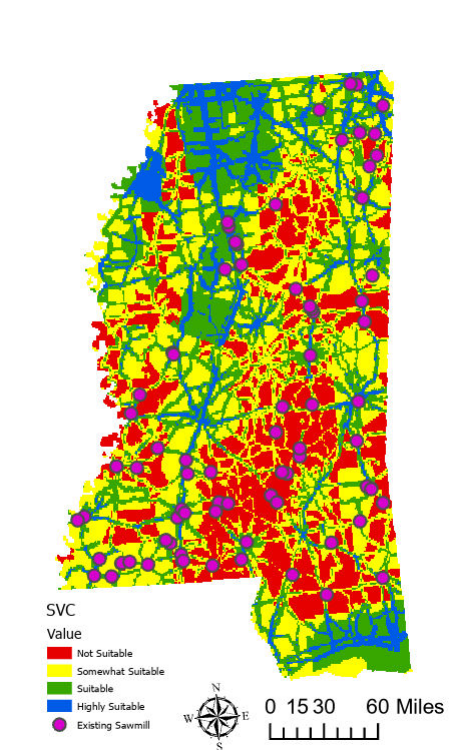}
    \includegraphics[width=0.32\linewidth]{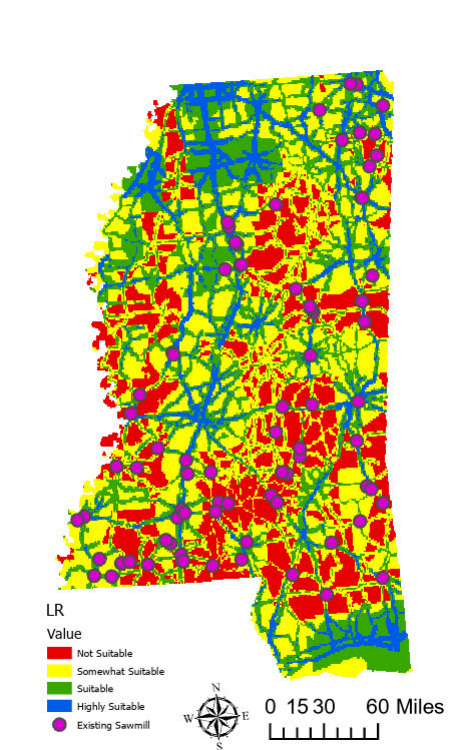}
    \includegraphics[width=0.32\linewidth]{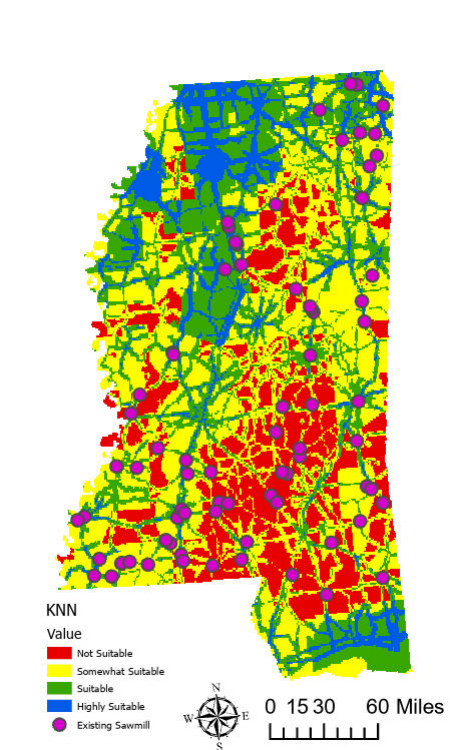}
    
    \caption{Suitability Map of Mississippi using five ML models}
    \label{Suitability_map}
\end{figure}

\begin{figure}[!ht]
    \centering
    \includegraphics[width=0.7\linewidth]{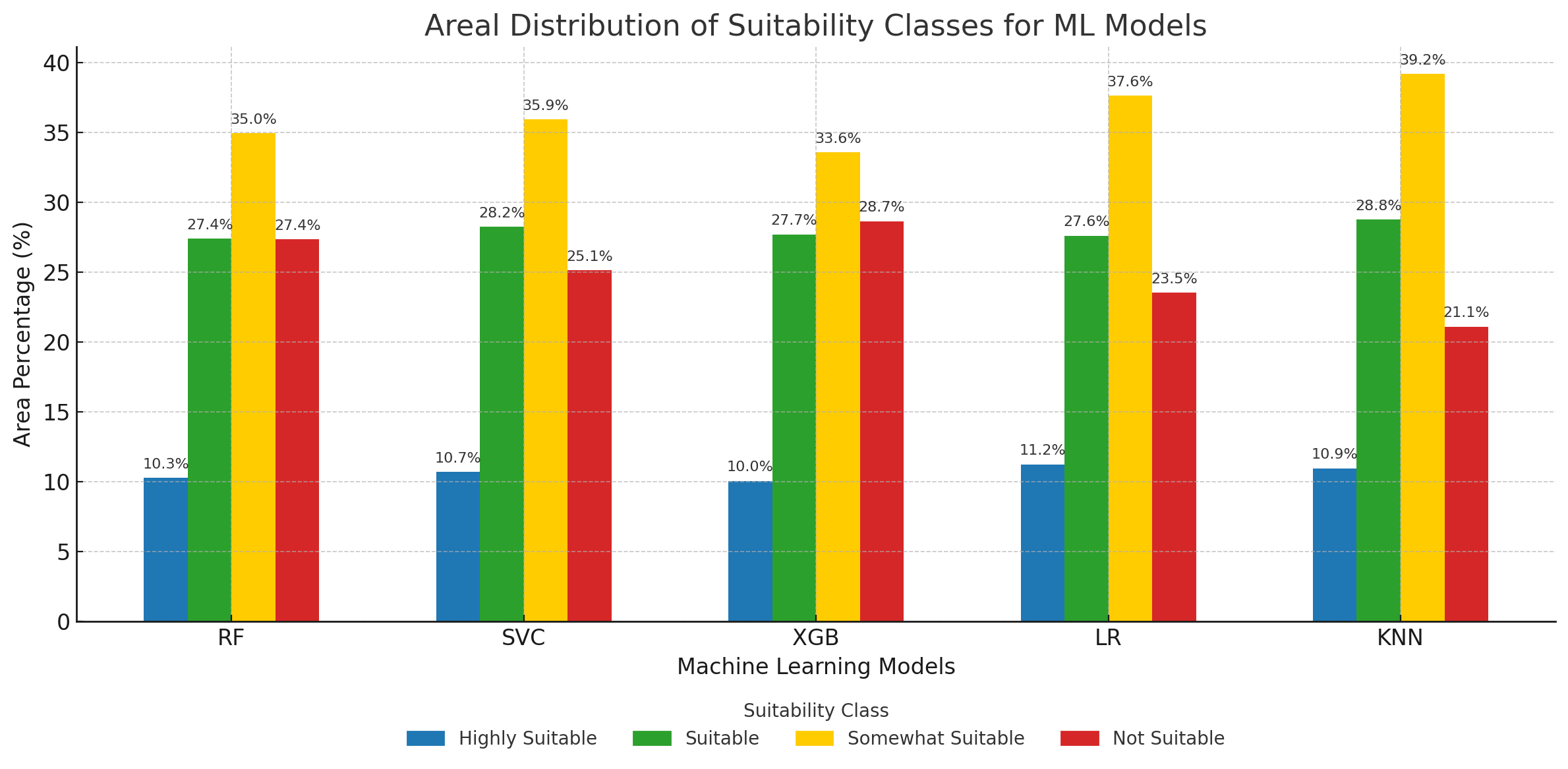}
    \caption{Area distribution of 4 Suitability Classes using RF, KNN, LR, XGB, and SVC model}
    \label{Areal_distribution}
\end{figure}

The site suitability maps generated using SHAP scores from each ML model are shown in Figure~\ref{Suitability_map}, while the corresponding areal distribution across suitability classes is depicted in Figure~\ref{Areal_distribution}. The results indicate that the majority of locations in Mississippi are classified as \textit{Somewhat Suitable}, with percentages ranging from 33.6\% to 39.3\% depending on the model. In contrast, the \textit{Highly Suitable} category consistently represents the smallest proportion of land area across all models. While the distribution patterns are broadly similar among the models, the KNN model predicts the largest combined percentage of area as either \textit{Highly Suitable} or \textit{Suitable}.

\begin{figure}[!ht]
    \centering
    \includegraphics[width=0.7\linewidth]{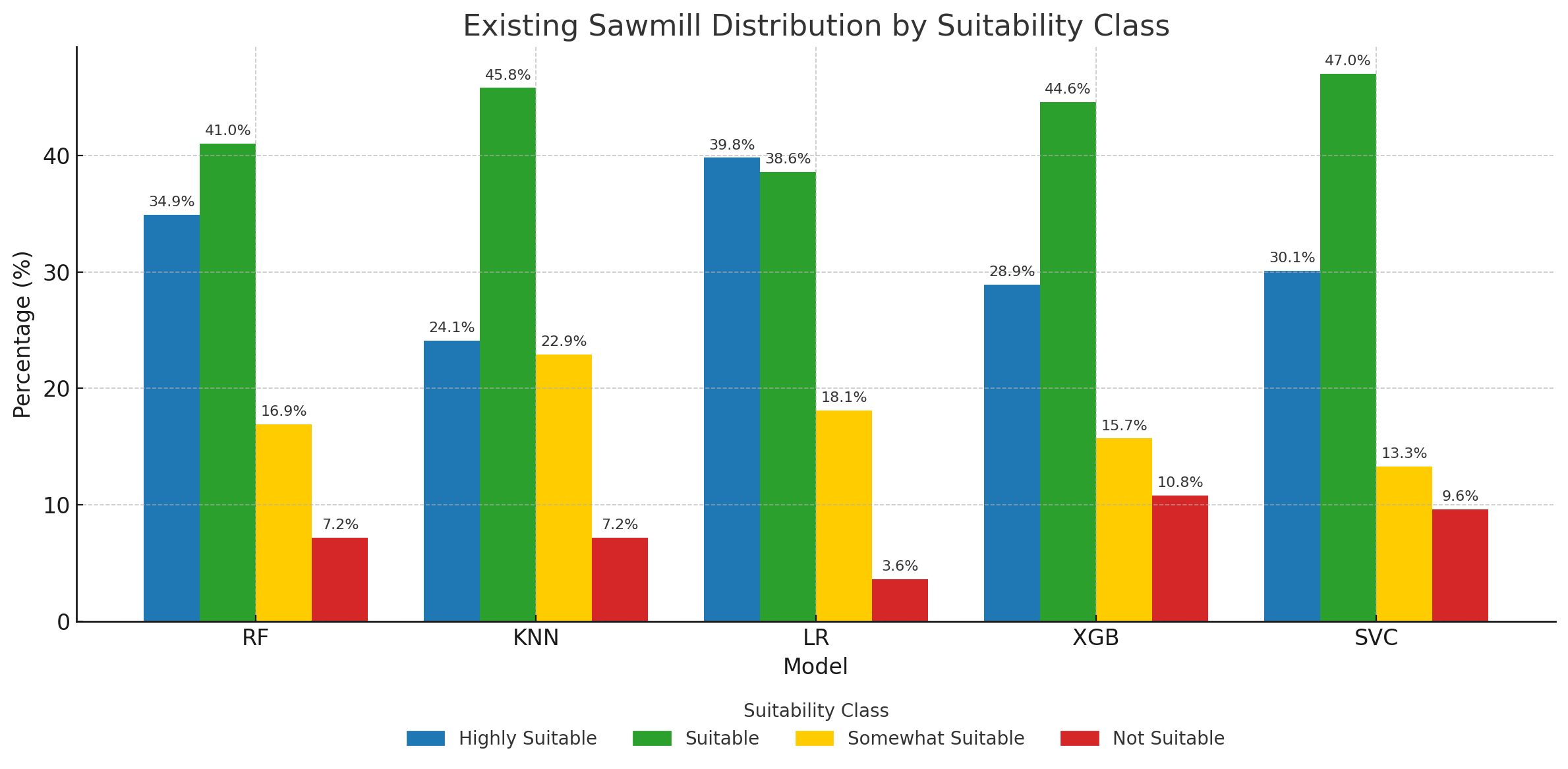}
    \caption{Existing Sawmill Distribution over the different Suitability Class}
    \label{Existing_sawmill}
\end{figure}

To validate the predictive accuracy of the generated suitability maps, we analyzed the spatial distribution of existing sawmills in Mississippi and calculated the percentage of sawmills located within each suitability class. This distribution is shown in Figure~\ref{Existing_sawmill}. As evident from the figure, the majority of existing sawmills are located in areas classified as either \textit{Highly Suitable} or \textit{Suitable}. Specifically, 74.9\% of sawmills fall within these two classes according to the RF model, while the corresponding values are 69.9\% for KNN, 78.4\% for LR, 71.5\% for XGB, and 77.1\% for SVC. Logistic Regression (LR) also reports the lowest percentage of sawmills in the \textit{Not Suitable} category (3.6\%), followed by 7.2\% for both RF and KNN, 10.8\% for XGB, and 9.6\% for SVC. We further validated the results through consultations with industry experts, who confirmed the model’s practical relevance and potential applicability in real-world decision-making. It is important to note that the dynamic nature of sawmill openings and closures continually reshapes market conditions; consequently, locations once considered highly suitable may become less suitable as new sawmills emerge nearby, intensifying market competition.

In addition to classification validation, candidate locations were rank-ordered based on their predicted likelihood scores. Each ML model produced continuous scores for all sites, which were sorted in descending order to create a prioritized list. This allowed for efficient filtering of high-ranking sites (i.e., top 10) for potential development.

\section{Discussion and Managerial Insights} \label{sec:Discussion_Insights}

Our case study offers several insights for decision-makers and researchers. 

First, the geography matters. The minimal relevance of \textit{Terrain Slope} and \textit{National Land Cover} on site suitability highlights the context-dependent nature of feature selection. In the case of Mississippi, the flat terrain and extensive forest coverage naturally reduce the relevance of these features. However, this finding does not necessarily generalize to other regions. In mountainous areas or regions with more diverse forest coverage, both \textit{Terrain Slope} and \textit{National Land Cover} could be more significant in terms of the suitability of candidate sawmill sites. 

Second, we find that market and supply-demand dynamics may be more important than proximity measures. Supply-demand ratio \textit{(SDR)}, a composite metric we proposed to serve a dual purpose in LB-MCDM framework: tracking local timber availability while simultaneously capturing existing local demand and competition, emerged as the most influential factor, regardless of the classification method used. Our proximity measures \textit{Road Distance, Rail Line Distance}, and \textit{Urban Area Distance} came second to \textit{SDR}, but are still essential by means of facilitating logistics and accessing utilities like power and water, and services like banking and healthcare that any sustainable operation needs.

Third, while the relatively low impact of \textit{Precipitation}, \textit{Market Revenue}, and \textit{Unemployment Rate} in this case study suggests that these factors may be of limited importance for sawmill site selection in Mississippi, it does not mean that they are universally unimportant. In regions where variability in precipitation significantly affects timber growth and harvesting or where economic factors such as market size, revenue, and workforce availability fluctuate considerably, these variables may play a more critical role. 

Fourth, transparency builds trust. One of the key strengths of the proposed framework is its enhanced explainability through SHAP analysis, which allows stakeholders to clearly understand the drivers behind site suitability. This transparency fosters stakeholder trust and supports more informed decision-making. Moreover, the model's validation against existing sawmill locations in Mississippi also demonstrates that it works in practice, not just in theory.

Fifth, the proposed method offers a practical tool for decision-makers to identify high-ranked locations based on predicted likelihood scores. Top-performing sites, such as the 10 highest-scoring or those with a score above 0.90, can be shortlisted for further evaluation. Then, decision-makers can combine these rankings with real-world considerations, such as land availability, infrastructure access, and alignment with regional development plans, to ensure practical site selection.

Sixth, the proposed framework can be adapted to other facility location problems beyond the timber industry, where multiple spatial and non-spatial criteria must be considered. The integration of ML, GIS, and MCDM can support decision-making in other sectors requiring location assessments. By reducing subjective biases and providing a systematic approach to evaluating candidate sites, this model serves as a practical decision-support tool for complex location planning challenges. 

Last but not least, the proposed framework can provide some relief for traditional facility location optimization models by processing large candidate sets to identify high-potential subsets, reducing problem size considerably and improving computational efficiency for NP-hard problems.

\section{Conclusion} \label{sec:Conclusion}

Our study contributes to the plant location literature in two ways: methodologically and practically. Our methodological contribution is the proposal of a novel methodology that addresses some of the limitations of existing methods. Traditional MCDM approaches rely heavily on subjective factor weighting, whereas conventional optimization and heuristic techniques mainly focus on proximity measures. The proposed LB-MCDM framework, on the other hand, integrates ML, MCDM, and GIS-based spatial analysis to objectively evaluate diverse spatial and non-spatial factors. Our ML models use multiple criteria derived from the literature and our conversations with the experts in the field, and generate \textit{a suitability map} that classifies all potential site locations into four categories as \textit{highly suitable, suitable, somewhat suitable}, and \textit{not suitable}. These categories are represented as color gradients on the GIS platform. Unlike traditional MCDM methods that begin with expert surveys to identify, measure, and rank multiple criteria, our approach starts with data and computation, generating unbiased and automated predictions with interpretable outputs to support expert decision-making. As the data changes, the suitability map is automatically updated. This approach is replicable and can be easily adapted to a wide range of industrial site location problems in other contexts.

In terms of practical contribution, we demonstrate the utility of the proposed model through a case study of the sawmill site location assessment problem in MS. Our computational results show that the RF model outperforms other algorithms, achieving an accuracy of 86.48\% and an AUC score of 0.9656. RF is followed by XGB (84.95\%, AUC = 0.9569) and SVC (81.86\%, AUC = 0.9159). LR and KNN demonstrated lower predictive performance. These models indicate that around 10–11\% of MS is \textit{highly suitable} for sawmill location, highlighting significant investment potential. The supply demand ratio \textit{(SDR)} stands out as the most influential factor emphasizing the importance of market dynamics. Proximity to roads, railways, and urban areas also plays a secondary role. We used multiple validation methods, including direct comparisons with past sawmill location decisions and expert opinions. The results demonstrate that our model's predictions align well with real-world data, as 70–80\% of existing sawmills are located in areas classified as either \textit{highly suitable} or \textit{suitable}. This validation gives us confidence that the LB-MCDM model offers a reliable framework for supporting data-driven industrial site assessments. 

This study has several limitations. We focused specifically on sawmill site selection in Mississippi, so some of our findings are inherently context-specific and cannot be generalized directly to other regions. For example, due to Mississippi’s flat and forest-rich landscape, \textit{terrain slope} and \textit{land cover} had minimal impact. These features would likely play a more crucial role in mountainous regions with larger variations in forest coverage. Furthermore, future research could examine other algorithms, such as artificial neural networks, to potentially enhance accuracy at the expense of reduced interpretability.

Future research could build on our work in several ways. Comparing multiple states or even countries with diverse geographic and socio-economic conditions would be valuable. Since our case study highlights how context shapes the role of specific features, it would be interesting to see how the importance of these factors (and potentially new ones) shifts across different settings. Additionally, incorporating other variables such as environmental regulations, land costs, and alternative transportation modes (i.e., water transportation) could add value. Last but not least, facility location optimization studies could use the proposed LB-MCDM framework to process large sets of candidate locations, filtering them down to a smaller subset of high-potential sites that meet multiple criteria. We expect that this would reduce both problem size and computational effort considerably.

\bibliographystyle{informs2014}

\newpage
\begin{APPENDIX}{}
\subsection*{A. Machine Learning Classifiers}
\subsubsection*{A.1 Random Forest Classifier}

The Random Forest (RF) classifier is an ensemble-based algorithm that aggregates the predictions of multiple decision trees to improve overall accuracy and reduce overfitting. Each tree \( T_1, T_2, \ldots, T_M \) is trained on a randomly selected subset of the training data and a random subset of features. This diversity among the trees helps capture various patterns in the data. The final prediction \( \hat{y} \) for an input feature vector \( \mathbf{x} \) is obtained through majority voting across all trees:
\begin{equation}
\hat{y} = \operatorname{mode} \left( T_1(\mathbf{x}), T_2(\mathbf{x}), \ldots, T_M(\mathbf{x}) \right)
\end{equation}
where \( M \) is the total number of trees in the forest, and \( T_m(\mathbf{x}) \) denotes the class prediction from the \( m \)-th decision tree. 


\subsubsection*{A.2 XGB Classifier}

XGBoost (XGB) is a gradient boosting framework that builds additive tree models iteratively. Each new tree \( g_m \) minimizes the regularized loss function:

\begin{equation}
\mathcal{J}(\theta) = \sum_{i=1}^{n} \ell(y_i, \hat{y}_i^{(m)}) + \sum_{m=1}^{M} \Omega(g_m)
\end{equation}

where \( \ell \) is the loss function between the true label \( y_i \) and the predicted value \( \hat{y}_i^{(m)} \), and \( \Omega \) represents the regularization to penalize complexity.

\subsubsection*{A.3 Support Vector Classifier}

Support Vector Classifier (SVC) constructs a decision boundary that maximizes the margin between classes. The prediction function \( \hat{y}_s \) is:

\begin{equation}
\hat{y}_s = \operatorname{sign}(\mathbf{a}^\top \mathbf{x} + c)
\end{equation}

where \( \mathbf{a} \) is the coefficient vector, \( c \) is the intercept, and \( \mathbf{x} \) is the input vector. We utilized the RBF kernel to handle non-linear separability.

\subsubsection*{A.4 Logistic Regression Classifier}

Logistic Regression (LR) is a linear classification algorithm that estimates class probabilities using the sigmoid activation function. For an input feature value \( i \), the predicted probability \( \hat{p} \) is computed as:

\begin{equation}
\hat{p} = \frac{1}{1 + \exp(- (w^\top i + q))}
\end{equation}

where \( w \) is the weight vector and \( q \) is the bias term. This function maps real-valued input into a probability between 0 and 1.


\subsubsection*{A.5 K-Nearest Neighbors Classifier}

The K-Nearest Neighbors (KNN) algorithm predicts the class \( \hat{y}_k \) of an input sample \( \mathbf{x} \) based on the majority vote of its \( k \)-nearest neighbors:

\begin{equation}
\hat{y}_k = \operatorname{mode}\left( y_j : \mathbf{x}_j \in \mathcal{N}_k(\mathbf{x}) \right)
\end{equation}

where \( \mathcal{N}_k(\mathbf{x}) \) is the set of the \( k \)-closest points to \( \mathbf{x} \) in the training set.


\subsection*{B. Performance Metrics}
The most critical aspect of evaluating machine learning models is the selection and interpretation of performance metrics. In this study, we used five evaluation metrics to compare model performance: Accuracy, Precision, Recall (Sensitivity), F1-Score, and Area Under the Curve (AUC). All of these metrics are derived from the confusion matrix, which consists of four essential components: True Positives (TP), False Positives (FP), True Negatives (TN), and False Negatives (FN). A True Positive (TP) refers to the number of correctly predicted positive instances. A False Positive (FP) represents the number of negative instances that were incorrectly predicted as positive. A True Negative (TN) is the number of correctly predicted negative instances, while a False Negative (FN) indicates the number of positive instances that were incorrectly predicted as negative.

Based on these components, the evaluation metrics are defined as follows:

\begin{align}
\text{Accuracy} &= \frac{TP + TN}{TP + TN + FP + FN} \\
\text{Precision} &= \frac{TP}{TP + FP} \\
\text{Recall} &= \frac{TP}{TP + FN} \\
\text{F1\text{-}Score} &= 2 \times \frac{\text{Precision} \times \text{Recall}}{\text{Precision} + \text{Recall}}
\end{align}

\textbf{Accuracy} evaluates the overall correctness of the model by measuring the proportion of total correct predictions. \textbf{Precision} focuses on the reliability of positive predictions, which is particularly critical in scenarios where false positives carry high costs. \textbf{Recall} measures the model’s ability to correctly identify actual positives, important when false negatives are especially undesirable. \textbf{F1-Score} is the harmonic mean of precision and recall, offering a balanced measure when dealing with imbalanced classes.

\textbf{AUC (Area Under the Curve)} evaluates a model’s ability to distinguish between classes by plotting the True Positive Rate (TPR) against the False Positive Rate (FPR) across all possible threshold values. While AUC is typically designed for binary classification tasks, our problem involves multiclass classification where AUC is not directly computed.

To adapt AUC for the multiclass context, we employed the One-vs-Rest (OvR) approach. In this method, AUC is calculated separately for each class by treating it as the positive class while combining all other classes as the negative class. This binary classification strategy is repeated for each class, and the resulting AUC scores are averaged to provide an overall measure of model performance in distinguishing each class from the rest.

Formally, for a multiclass problem with $K$ classes, the OvR AUC is defined as:

\begin{equation}
\text{AUC}_{\text{OvR}} = \frac{1}{K} \sum_{k=1}^{K} \text{AUC}(y_k, \hat{p}_k)
\end{equation}

where:
\begin{itemize}
\item $y_k$ is the binary indicator (1 or 0) for whether the true class is $k$,
\item $\hat{p}_k$ is the predicted probability for class $k$,
\item $\text{AUC}(y_k, \hat{p}_k)$ is the AUC score for class $k$ versus all others.
\end{itemize}

Together, these metrics provide a comprehensive understanding of the model’s performance, ensuring robust and reliable evaluation under various data conditions.

\subsection*{C. Feature Importance and Explainability} 
Understanding the rationale behind an ML model’s prediction by identifying how input features contribute to the output is a crucial aspect of predictive analytics and model interpretability. Traditional ML models often rely on built-in feature selection methods or importance metrics such as information gain (IG) to quantify feature relevance. However, as \cite{lundberg2018consistent} points out, gain-based feature importance measures can be inconsistent and misleading. To overcome these limitations, SHapley Additive exPlanations (SHAP) has emerged as a state-of-the-art local explainability technique. SHAP is grounded in cooperative game theory, originally introduced by Shapley \cite{shapley1953stochastic}, and attributes the prediction of a model to each feature by fairly distributing the output among the contributing features.

The SHAP value for a feature \( i \) is defined as:

\begin{equation}
\phi_i = \sum_{S \subseteq N \setminus \{i\}} \frac{|S|! (|N| - |S| - 1)!}{|N|!} \left[ f(S \cup \{i\}) - f(S) \right]
\end{equation}

where:
\begin{itemize}
  \item \( N \) is the set of all features,
  \item \( S \) is a subset of features not containing \( i \),
  \item \( f(S) \) is the model prediction based only on the features in subset \( S \),
  \item \( \phi_i \) is the SHAP value for feature \( i \), representing its marginal contribution.
\end{itemize}

This formulation allows SHAP to estimate the positive or negative contribution of each feature to a given prediction (local interpretability), while also aggregating these contributions to understand the feature's overall importance across all predictions (global interpretability).

\end{APPENDIX}

\end{document}